\documentclass[runningheads]{llncs}
\usepackage[T1]{fontenc}
\usepackage{graphicx}
\usepackage{booktabs}
\usepackage[misc]{ifsym}
\newcommand{\corr}{(\Letter)}

\usepackage{algorithm}
\usepackage{algpseudocode}
\usepackage{amsmath}   
\usepackage{amssymb}   
\usepackage{amsfonts}  
\usepackage{array}     
\usepackage{multirow}  
\usepackage{tabularx}
\usepackage{makecell}
\usepackage[table]{xcolor} 
\usepackage{cite}      

\begin{document}

\title{Modeling Inverse Ellipsometry Problem via Flow Matching with a Large-Scale Dataset}

\titlerunning{EllipBench}

\author{Yiming Ma\inst{1}\thanks{Equal Contribution.} \and
Jianzhi Teng\inst{2}\protect\footnotemark[1] \and
Xinjie Li\inst{3}\protect\footnotemark[1] \and
Xin Sun\inst{4} \and
Zhiyong Wang\inst{4} \and
Yuzhou Song\inst{5} \and
Lionel Z. Wang\inst{2}\textsuperscript{, \corr} \and
Bin Chen\inst{1,6}\textsuperscript{, \corr}}

\authorrunning{Y. Ma et al.}

\institute{Chongqing Research Institute of Harbin Institute of Technology \and
The Hong Kong Polytechnic University \and
The Pennsylvania State University \and
Tianjin University \and
Tsinghua University \and
Harbin Institute of Technology (Shenzhen)}


\begingroup
\renewcommand\thefootnote{}
\footnotetext{\raggedright \corr~Corresponding authors: \texttt{zhe-leo.wang@connect.polyu.hk}, \texttt{chenbin2020@hit.edu.cn}}
\endgroup

\maketitle              

\begin{abstract}
Inverse ellipsometry, i.e., reconstructing optical constants and film thickness from the measured phase difference $\Delta$ and amplitude ratio $\Psi$, is a fundamentally ill-posed problem. Traditional solutions rely on slow, expert-driven iterative fitting, while the development of machine learning approaches has been severely limited by the lack of large-scale, physically consistent datasets. To address this gap, we introduce \textbf{EllipBench}, a comprehensive benchmark comprising over 8 million high-precision samples spanning 98 thin-film materials and 5 substrates. Building upon this benchmark, we conduct a systematic evaluation of a broad spectrum of methods, including traditional machine learning models, deep neural networks, and Physics-Informed Neural Networks, and show that existing paradigms consistently struggle to fully resolve the inverse ellipsometry task. To better capture its inherent ambiguity, we further propose a novel \textbf{Decoupled Conditional Flow Matching (DCFM)} framework. Rather than formulating the problem as deterministic point-to-point regression, DCFM explicitly decouples geometric film thickness and incorporates it as a robust physical condition to guide a continuous vector field for modeling the inverse probability distribution of wavelength-dependent optical constants. Combined with a gradient detachment strategy and physics-based constraints, our joint architecture effectively mitigates intrinsic physical ambiguities and delivers a robust and accurate solution for inverse ellipsometry.

\end{abstract}

\section{Introduction}
\begin{figure}[t]
\begin{center}
\centerline{\includegraphics[width=0.8\columnwidth]{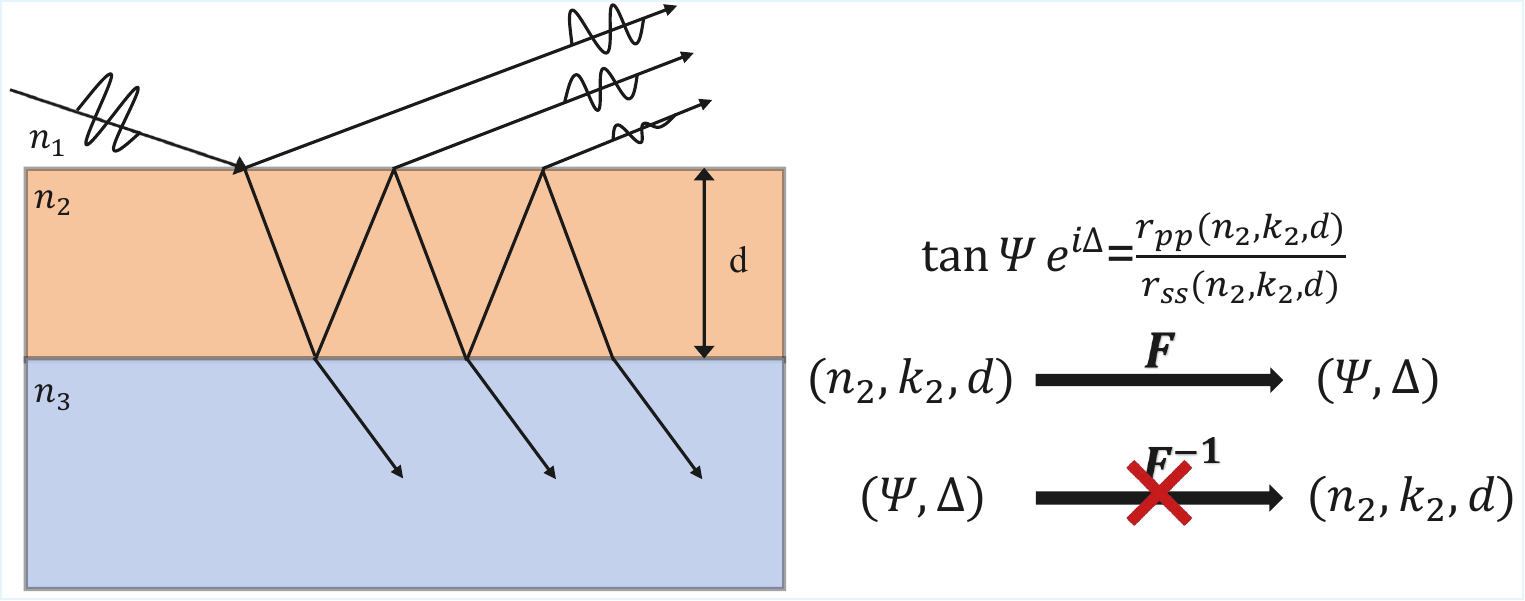}}
\caption{Schematic of light refraction and interference in a thin film system. The forward process analytically maps the unknown film properties ($n_2$, $k_2$, $d$) to the measured ellipsometric parameters $\Psi$ and $\Delta$. However, the inverse mapping lacks an exact analytical formula.}
\label{fig:refraction}
\end{center}
\end{figure}

Thin films are essential components in modern technological applications, including optoelectronics, microelectronics, energy conversion, and aerospace engineering \cite{macleod2010thin}. Their optical properties, specifically the refractive index and extinction coefficient, heavily dictate the performance of these advanced devices \cite{poelman2003methods}. Spectroscopic ellipsometry serves as a fundamental optical metrology technique for characterizing these properties alongside film thickness \cite{fujiwara2007spectroscopic}. The core principle of ellipsometry is illustrated in Figure \ref{fig:refraction}. This technique offers significant advantages over alternative analytical tools. It is non-destructive, requires no specialized sample preparation, and provides exceptional precision for ultra-thin films down to the sub-nanometer scale, outperforming traditional interferometry by orders of magnitude \cite{tompkins2005handbook}.

Despite these advantages, ellipsometry is fundamentally an indirect measurement technique. Determining the physical parameters requires complex mathematical modeling to map the measured amplitude ratio $\Psi$ and phase difference $\Delta$ back to the optical constants and thickness \cite{collins2000recent}. This classical inversion relies heavily on iterative regression algorithms, such as the Levenberg-Marquardt method, to minimize the discrepancy between experimental data and theoretical models \cite{johs1993regression}. As shown in Figure \ref{fig:refraction}, finding an exact analytical solution $F^{-1}$ for the inverse ellipsometry problem is generally unfeasible. The conventional fitting procedures demand extensive expert intervention to establish accurate optical models and provide reasonable initial guesses for the sample properties \cite{zhu2024lightweight}. Without proper initialization, algorithms frequently converge to non-physical local minima. This reliance on human expertise makes the analysis exceptionally slow and difficult to scale for high-throughput applications. Furthermore, the inverse mapping is notoriously ill-posed. Different combinations of film thickness and optical constants can produce nearly identical ellipsometric spectra, leading to severe one-to-many mapping ambiguities \cite{adler2017solving}. Developing computational frameworks to robustly reconstruct these parameters without manual intervention represents a critical challenge in optical metrology \cite{oates2011characterization}.

Machine learning has recently shown strong potential in materials science and physical property characterization \cite{wei2019machine}. These data-driven methodologies offer a promising pathway to bypass traditional iterative fitting. However, the development of robust large-scale models for ellipsometry has been severely restricted by the lack of massive, publicly available, and physically consistent data. To bridge this gap, we construct a comprehensive large-scale ellipsometry dataset, EllipBench, which is dedicated to thin film optical properties. This dataset encompasses 98 distinct thin film materials deposited on 5 different substrates, yielding over 8 million highly precise data points. Detailed specifications are provided in Table \ref{tab1:dataset}. To systematically evaluate computational capabilities on this complex task, we test a wide range of methods. These encompass traditional machine learning algorithms as well as Physics-Informed Neural Networks (PINNs) that integrate fundamental optical equations into the optimization process \cite{karniadakis2021physics,raissi2019physics}. While these methods demonstrate varying degrees of success, they struggle to fully resolve the inherent ambiguities of the ellipsometric inversion process.


To better address the inverse ellipsometry problem, we propose a novel Decoupled Conditional Flow Matching framework \cite{albergo2023building,lipmanflow}. Instead of relying on deterministic point-to-point regression, our method decouples geometric thickness from wavelength-dependent optical constants and uses thickness as a physical condition to guide the modeling of their inverse distribution \cite{liu2018training}. Combined with physics-based constraints, this design provides a robust and effective solution for inverse ellipsometry.

\begin{table*}[t]
\centering
\caption{Film and substrate materials in the EllipBench.}
\label{tab1:dataset}
\small
\renewcommand{\arraystretch}{1.3}
\resizebox{\textwidth}{!}{%
\begin{tabular}{@{} l l >{\raggedright\arraybackslash}p{0.72\textwidth} @{}}
\toprule
\textbf{Type} & \textbf{Category} & \textbf{Materials} \\
\midrule
\multirow{5}{*}{Film} & Metal & Ag, Al, Au, Be, Cd, Co, Cr, Cs, Cu, Fe, Hg, In, Ir, K, Li, Mg, Mn, Mo, Na, Nb, Ni, Os, Pd, Pt, Re, Rh, Sn, Ta, Ti, V, W, Zr, Ru \\
\cmidrule{2-3}
& Alloy & AlTi, AlTiC, In$_{53}$GaAs, NiFe, NiP, TiW \\
\cmidrule{2-3}
& Compound & Al$_2$O$_3$, AlAs, AlGaAs, AlN, AlSb, BaTiO$_3$, CaF$_2$, CdS, CdSe, CdTe, Cr$_2$O$_3$, Cu$_2$O, Fe$_2$O$_3$, GaAs, GaN, GaP, GaSb, CuO, HgSe, HgTe, In$_2$O$_3$, InAs, InGaAs, InN, InP, InSb, ITO, MgF$_2$, MgO, Mn$_2$O$_3$, Nb$_2$O$_5$, PbS, PbSe, PbTe, Si$_3$N$_4$, SiC, SiO, SiO$_2$, Ta$_2$O$_5$, Ta$_3$N$_5$, TaN, ZrO$_2$, TiC, TiN, TiO$_2$, TiSi, VN, ZnS, ZnSe, ZrN \\
\cmidrule{2-3}
& Polymer & PDMS, PMMA \\
\cmidrule{2-3}
& Others & Glass, Diamond, Si, Ge, Te, Sb \\
\midrule
Substrate & Others & a-Si, ITO, SrTiO$_3$, Si, PI \\
\bottomrule
\end{tabular}%
}
\end{table*}
\subsection{Main Contributions}
\textbf{First}, we introduce an open-source, large-scale inverse ellipsometry dataset EllipBench, accompanied by comprehensive baseline benchmarking. \textbf{Second}, we introduce EC Error, a physics-inspired residual metric for analyzing power-balance consistency and identifying physically challenging regions in the benchmark. \textbf{Finally}, we present Decoupled Conditional Flow Matching (DCFM), a novel framework that achieves state-of-the-art performance on our benchmark.

\section{Related Work}

\subsection{Ellipsometry Datasets}
The progression of computational methods in optical metrology heavily relies on the availability of comprehensive datasets. However, the ellipsometry domain currently lacks large-scale and publicly accessible benchmarks. Previous studies often relied on private or simulated data collections. For instance, Jiang et al. \cite{jiang2024generic} utilized a dataset of 660000 simulated grating parameters for deep learning assisted analysis. Similarly, Wang et al. \cite{wang2023ellipsonet} generated 90000 input output pairs derived from 450000 multilayer stack structures using the C2DB database for their EllipsoNet model, and subsequently applied a comparable dataset for ReflectoNet \cite{wang2023measuring}. Arunachalam et al. \cite{arunachalam2022machine} employed a mixture of experimental and simulated titanium dioxide data without specifying the exact volume. Liu et al. \cite{liu2021machine} utilized a much smaller collection comprising 6240 parameter pairs sourced from the established Palik and Sopra databases. The restricted scale and private nature of these datasets limit the development and rigorous evaluation of advanced algorithms. In contrast, our work introduces a publicly reproducible dataset exceeding 8 million entries across 98 materials, establishing a robust foundation for future research.

\subsection{Mathematics-Based Inversion Methods}
Ellipsometry is an indispensable optical measurement technique in semiconductor and optoelectronic manufacturing \cite{fujiwara2007spectroscopic}. As an indirect measurement, extracting physical properties (e.g., optical constants and film thickness) from the acquired amplitude ratio $\Psi$ and phase difference $\Delta$ constitutes a classic, intrinsically ill-posed inverse problem \cite{bell1978solutions, akbalik2009inverse}. Traditionally, resolving this requires constructing a layered physical model via Fresnel equations \cite{tompkins2005handbook} and employing iterative optimization (e.g., Levenberg-Marquardt) to minimize the discrepancy between theoretical and experimental spectra \cite{johs1993regression,losurdo2013ellipsometry}. However, this fitting procedure is computationally demanding due to severe parameter correlations and frequent convergence to non-physical local minima \cite{jellison1998spectroscopic}. Consequently, it relies heavily on expert intervention for model definition and parameter initialization, rendering it highly inefficient for modern high-throughput applications \cite{zhu2024lightweight}.

\subsection{Machine-Learning-Based Inversion Methods}
The rapid evolution of machine learning has introduced powerful large-scale models to tackle complex scientific problems, achieving remarkable success across diverse disciplines such as structural biology, computational chemistry, materials discovery, and quantum physics \cite{jumper2021highly,stokes2020deep,merchant2023scaling,carleo2019machine}. Inspired by these achievements, researchers have increasingly integrated deep learning architectures into ellipsometric data analysis to circumvent the limitations of traditional mathematical fitting. Early attempts by Urban et al. \cite{urban1994real} utilized shallow neural networks as preprocessors to generate better initial parameter guesses for subsequent regression algorithms. More recently, data-driven approaches have attempted direct parameter prediction. As discussed previously, models like EllipsoNet \cite{wang2023ellipsonet} and ReflectoNet \cite{wang2023measuring} utilize deep neural networks to approximate the inverse mapping. Other works have explored PINNs to embed physical constraints directly into the learning process \cite{karniadakis2021physics}. Despite these efforts, conventional regression networks struggle significantly with the extreme one-to-many mapping ambiguity inherent in inverse ellipsometry. In these ill-posed scenarios, identical optical spectra can originate from entirely different physical configurations, causing standard deterministic models to predict physically invalid averages of multiple possible states \cite{ardizzone2018analyzing}. To resolve this mathematical ambiguity, recent advancements in continuous normalizing flows and probabilistic modeling have demonstrated immense potential in capturing complex posterior distributions for physical inverse problems \cite{kobyzev2020normalizing}. Building upon these theoretical foundations, our work diverges entirely from standard deterministic regression. We propose the Decoupled Conditional Flow Matching framework. 


\section{EllipBench}
\label{benchmark}

\subsection{Task Statement and Problem Formulation}
\label{task statement}

As shown in Figure \ref{fig:refraction}, the forward process of ellipsometry involves calculating the ellipsometric parameters $\Delta$ and $\Psi$ from the physical properties of the materials. This forward mapping is governed by fundamental optical interactions, beginning with Snell's law. Let $N = n + i k$ represent the complex refractive index and $\theta$ represent the angle of refraction. The relationship across the air (1), thin film (2), and substrate (3) interfaces is defined as:
\begin{equation}
    N_1 \sin \theta_1 = N_2 \sin \theta_2 = N_3 \sin \theta_3
\end{equation}

Based on these refraction angles, the Fresnel equations characterize the reflection behavior at each interface \cite{azzam1978ellipsometry}. These single layer reflections are subsequently extended to multi-layer structures to derive the total complex amplitude reflection coefficients $r_{pp}$ and $r_{ss}$ for parallel and perpendicular polarization states. The detailed mathematical derivations for these intermediate reflection coefficients and their corresponding phase differences are provided in Appendix. Ultimately, this forward model establishes the precise mathematical mapping from the physical properties of the thin films to the measured ellipsometric spectra:
\begin{equation}
    \tan \Psi \exp(i \Delta) = \frac{r_{pp}}{r_{ss}}
\end{equation}

In optical metrology, the practical objective is to solve the inverse ellipsometry problem. This task aims to deduce the unknown film thickness $d$ and optical constants $n_2$ and $k_2$ directly from the measured optical spectra $\Delta$ and $\Psi$. This inversion is an inherently ill-posed problem \cite{tompkins2005handbook}. The inverse mapping is highly non-linear and suffers from severe parameter correlations \cite{jellison1993data}. Most critically, it presents an extreme one-to-many mapping ambiguity where completely different physical configurations of film thickness and optical constants can produce nearly identical ellipsometric spectra \cite{fujiwara2007spectroscopic}. 

When modeling this inverse problem for deep learning approaches, the objective is to approximate this inverse mapping function \cite{ongie2020deep}. The input features to the neural network inherently consist of the measured ellipsometry angles $\Psi$ and $\Delta$, the wavelength of the incident light $\lambda$, and the known optical constants of the substrate $n_3$ and $k_3$. The target outputs that the network must predict are the corresponding properties of the thin film, specifically $n_2$, $k_2$, and $d$. 


\subsection{Dataset Physical Consistency Validation}

To assess the physical consistency of the generated dataset, we introduce the Energy Conservation Error (EC Error) as a practical quality assessment metric. The metric is derived from the optical power-balance relation and measures the residual deviation of the reflected and transmitted energy from unity for the two polarization states \cite{macleod2010thin}. We define it as
\begin{equation}
\text{EC Error} = \frac{1}{N}\sum_{i=1}^{N}\left(|R_{p,i} + T_{p,i} - 1| + |R_{s,i} + T_{s,i} - 1|\right),
\end{equation}
where $N$ denotes the total number of samples. Here, the reflectance terms are computed from the complex reflection coefficients as $R_p = |r_{123,p}|^2$ and $R_s = |r_{123,s}|^2$, while the transmittance terms are computed from the complex transmission coefficients as
$T_p = \frac{n_3\cos\theta_3}{n_1\cos\theta_1}|t_{123,p}|^2$ and
$T_s = \frac{n_3\cos\theta_3}{n_1\cos\theta_1}|t_{123,s}|^2$.
The coefficients $r_{123,\cdot}$ and $t_{123,\cdot}$ are obtained from the standard multilayer interference formalism.

Since EllipBench includes absorptive materials, $R+T$ is not expected to equal $1$ exactly in general. Therefore, EC Error should be interpreted as a residual-based consistency indicator rather than a strict violation measure. Smaller values indicate closer agreement with the idealized power-balance condition, whereas larger values typically highlight strongly absorptive or numerically sensitive regimes. In this work, we use EC Error as a lightweight physical diagnostic to profile data quality across material categories, thickness ranges, and wavelength intervals. Detailed derivations and implementation procedures are provided in the Appendix.

\begin{figure}[t]
\centering
\includegraphics[width=\columnwidth]{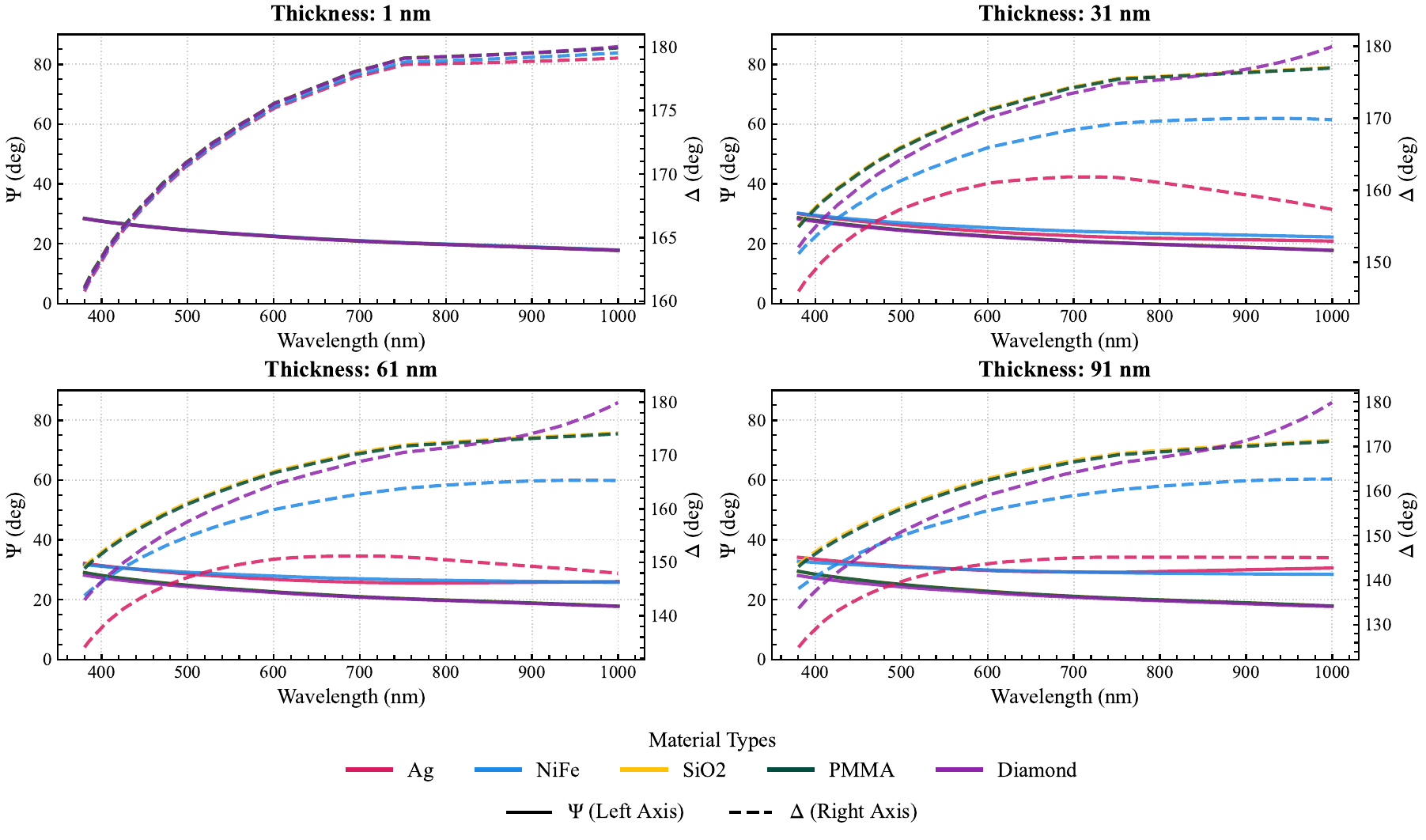}
\caption{Simulated ellipsometric spectra for five representative thin films on an amorphous silicon substrate across four thickness levels. Solid and dashed lines denote the amplitude ratio $\Psi$ on the left axis and phase difference $\Delta$ on the right axis over the 380 to 1000 nm wavelength range.}
\label{fig:spectra}
\end{figure}

\subsection{Dataset Generation and Characteristics}
We introduce EllipBench, a comprehensive benchmark designed for inverse ellipsometry research utilizing machine-learning methods. The dataset builds upon experimentally derived optical constants for 98 thin film materials (metals, alloys, inorganic compounds, and polymers) across five widely used substrates (amorphous Silicon, Indium Tin Oxide, crystalline silicon, Strontium Titanate, and Polyimide). Spectral data were sampled from 380.28 to 999.87 nm with a 2.6 nm resolution. We generated ellipsometric spectra via the forward Fresnel formalism at a 70 degree incident angle, systematically varying film thickness across 20 logarithmic steps from 1 to 96 nm. After applying a strict energy conservation filter to guarantee physical validity \cite{lucarini2005kramers}, the final dataset comprises over 8 million entries.

To illustrate the inherent physical complexity, Figure \ref{fig:spectra} presents the spectral variations for five representative materials on amorphous Silicon. As film thickness increases, the optical responses exhibit highly non-linear behaviors and pronounced interference oscillations. The intricate crossing of these curves visually demonstrates the extreme one-to-many mapping ambiguity, reinforcing the necessity of a massive dataset to train robust analytical models.

\begin{figure}[t]
  \centering
  \includegraphics[width=\columnwidth]{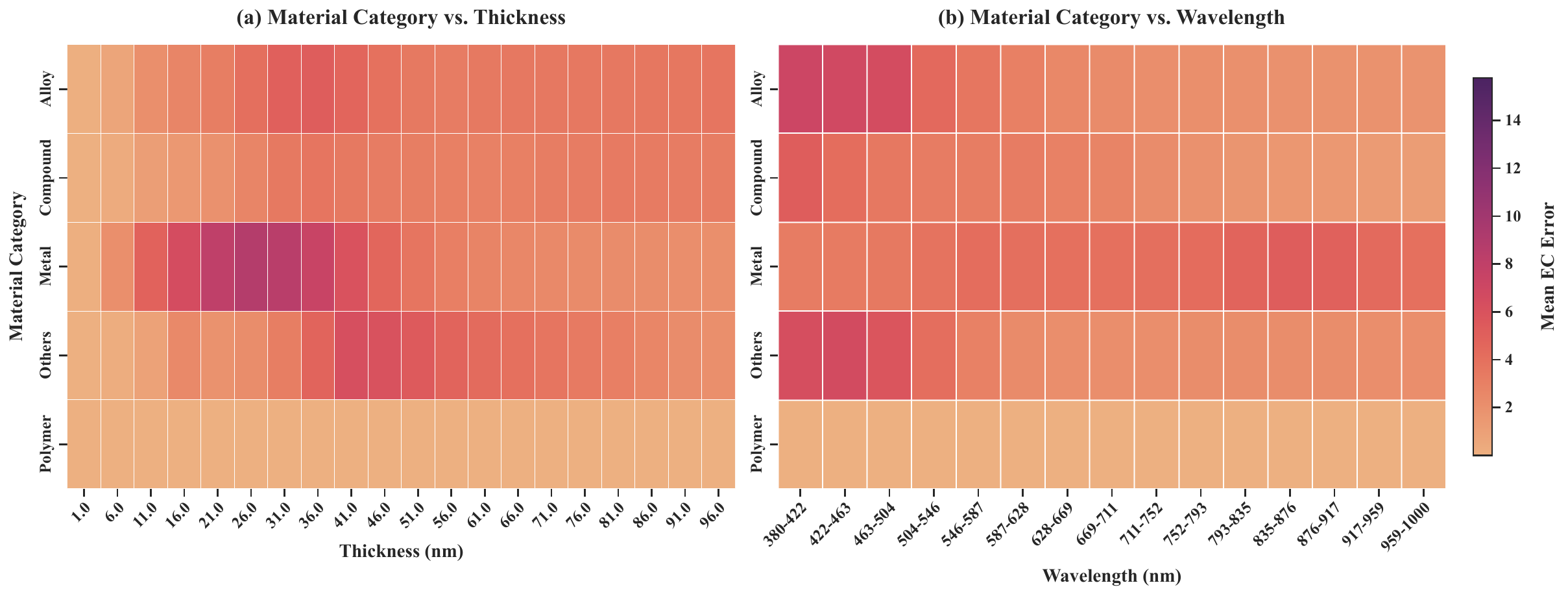}
  \caption{Energy Conservation Error Analysis across material categories: (a) Error distribution across thickness; (b) Error distribution across wavelength spectral ranges.}
  \label{fig:combined_ec_error}
\end{figure}

To further investigate data quality, Figure \ref{fig:combined_ec_error} visualizes the energy conservation error across thickness and wavelength domains. The error exhibits a non-linear correlation with film thickness, with the Metal category presenting significant measurement challenges in the 16 to 31 nm range. This indicates a potential blind spot for highly reflective materials caused by destructive interference and surface–bulk interactions. Conversely, Polymer and Compound categories maintain consistently low error levels. Spectral analysis further reveals that metallic films exhibit pronounced error intensities in the 380 to 600 nm region. Ultimately, EllipBench effectively captures complex thickness dependent optical behaviors, but researchers must carefully consider these inherent physical measurement limitations when utilizing the dataset.

\section{Method}
\label{method}

In this section, we present our complete approach for solving the ill-posed inverse ellipsometry problem \cite{fujiwara2007spectroscopic}. Extracting physical parameters from measured optical spectra requires overcoming severe one-to-many mapping ambiguities. Traditional regression networks often fail in this context by predicting the non-physical mean of multiple valid states. To overcome this fundamental limitation, we propose the Decoupled Conditional Flow Matching framework. Instead of performing deterministic point regression, our method constructs a continuous probability density path to explicitly model the complex posterior distribution of the physical properties, conditioned on rigorously decoupled geometric constraints.

\subsection{Preliminaries on Rectified Flow Matching}
Flow Matching trains continuous normalizing flows to transform a base probability distribution into a complex target distribution via an ordinary differential equation \cite{lipmanflow}. For target physical parameters $\tilde{\mathbf{y}} \sim p_{data}(\tilde{\mathbf{y}})$ and base Gaussian noise $\epsilon \sim \mathcal{N}(0, I)$, Rectified Flow Matching constructs a linear interpolation path over time $t \in [0, 1]$:
\begin{equation}
    \tilde{\mathbf{y}}_t = (1 - t) \epsilon + t \tilde{\mathbf{y}}
\end{equation}
A neural vector field $v_\theta$ is optimized via mean squared error to approximate the target velocity $\mathbf{u}^\star = \tilde{\mathbf{y}} - \epsilon$ driving this transition. During inference, numerical integration solves the differential equation forward from $t=0$ to $t=1$, reconstructing the physical parameters from pure noise \cite{chen2018neural}.

\subsection{Decoupled Conditional Flow Matching Framework}

\begin{figure}[t]
  \centering
  \includegraphics[width=\columnwidth]{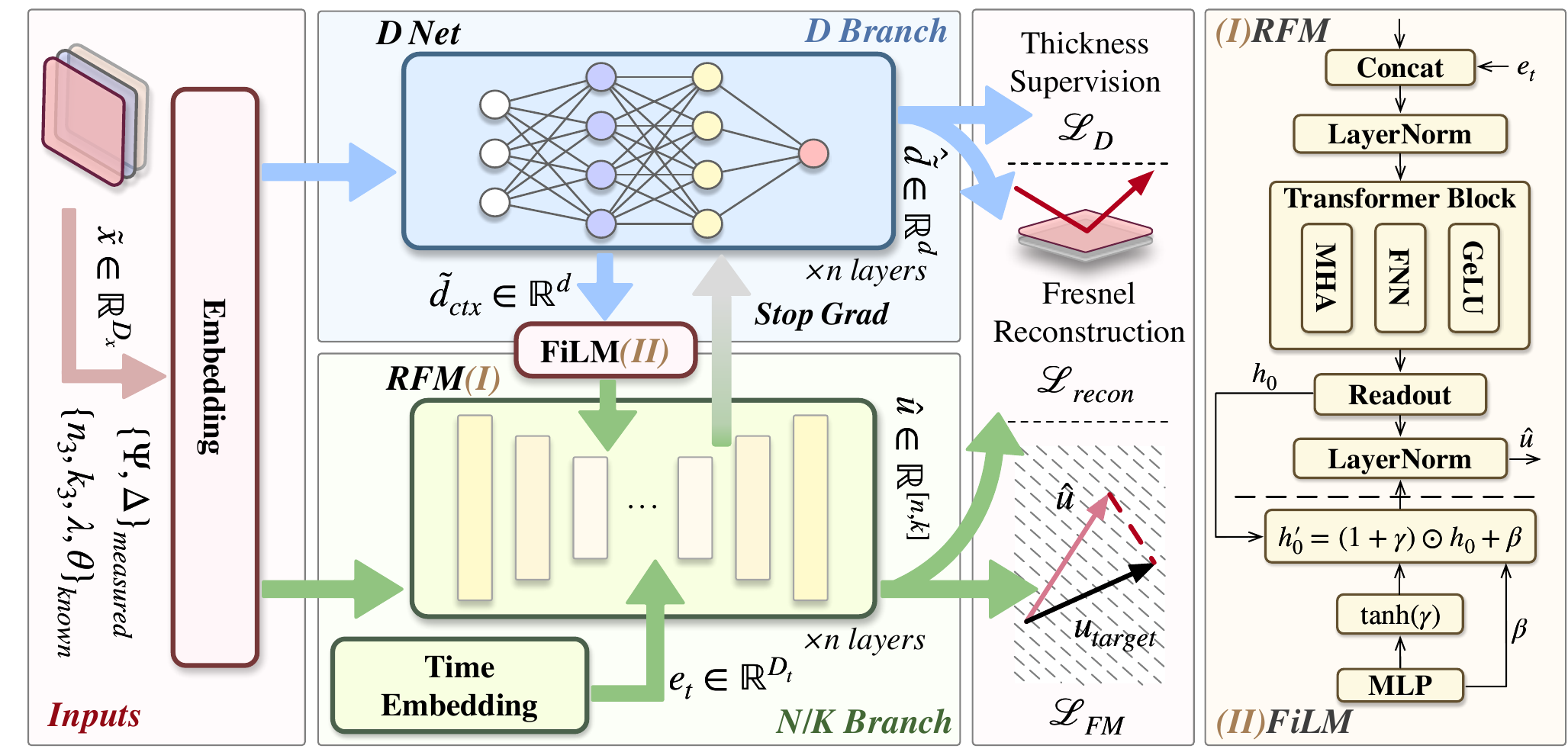}
  \caption{Overview of the Decoupled Conditional Flow Matching framework. A deterministic predictor extracts the geometric thickness, injecting it via Feature-wise Linear Modulation into a conditional vector field network that models the optical constants. Gradient detachment isolates the thickness branch for training stability. The system is jointly optimized using thickness supervision, vector field matching, and an optional physics-based Fresnel reconstruction loss.}
  \label{fig:framework}
\end{figure}

Solving inverse optical problems requires strict conditioning on measured spectra. Furthermore, directly predicting the physically entangled parameters of geometric film thickness and wavelength dependent optical constants inevitably destabilizes generative training. To alleviate this bottleneck, our framework, as illustrated in Figure \ref{fig:framework} and formalized in Algorithm \ref{alg:point-fm-transformer}, explicitly isolates the macroscopic geometric thickness from the microscopic continuous dispersion properties. We process these modalities through separate specialized branches while jointly optimizing them in a single backward pass. Let the input point-wise spectral features be denoted as $\tilde{\mathbf{x}} \in \mathbb{R}^{D_x}$, the target optical constants as $\tilde{\mathbf{y}} = [\tilde{n}, \tilde{k}] \in \mathbb{R}^2$, and the ground truth thickness as $\tilde{d}$.

\begin{algorithm}[t]
\caption{Conditional Rectified Flow Matching}
\label{alg:point-fm-transformer}
\begin{algorithmic}[1]
\State \textbf{Input:} point dataset \(\mathcal{D}=\{(\tilde{x}_i,\tilde{y}_i,\tilde{d}_i)\}\), vector field \(v_\theta\), optional thickness context \(\tilde{d}_{\mathrm{ctx}}\), time-embed dim \(D_t\)
\For{iteration \(=1,2,\dots\)}
    \State Sample minibatch \((\tilde{x},\tilde{y},\tilde{d})\sim\mathcal{D}\)

    \State Construct thickness context \(\tilde{d}_{\mathrm{ctx}}\)
    \Comment{\(\tilde{d}_{\mathrm{ctx}}\in\{\varnothing,\tilde{d},\mathrm{stopgrad}(\hat{\tilde{d}})\}\)}

    \State Sample \(t\sim\mathcal{U}(0,1)\), \(\epsilon\sim\mathcal{N}(0,I)\)
    \Comment{\(t\in[0,1]\), \(\epsilon\in\mathbb{R}^{y_{\mathrm{dim}}}\)}

    \State \(\tilde{y}_t \gets (1-t)\epsilon + t\tilde{y}\)
    \Comment{Rectified interpolation }

    \State \(u^\star \gets \tilde{y}-\epsilon\)
    \Comment{Target velocity field }

    \State \(h_0 \gets \mathrm{TransformerEnc}([\mathrm{tok}_{\tilde{y}},\mathrm{tok}_{\tilde{x}},\mathrm{tok}_t])\)
    \Comment{readout from \(\mathrm{tok}_{\tilde{y}}\)}
    
    \State \(\hat{u} \gets \mathrm{FiLM}(h_0,\tilde{d}_{\mathrm{ctx}})\) \hfill 
    \Comment{FiLM-modulated readout}

    \State \(\mathcal{L}_{\mathrm{FM}} \gets \lVert \hat{u}-u^\star\rVert_2^2\)

    \State Update(\(\theta\), \(\nabla_\theta \mathcal{L}_{\mathrm{FM}}\))
    \Comment{gradient step}
\EndFor
\State \Return \(v_\theta\)
\end{algorithmic}
\end{algorithm}

\subsubsection{Decoupled Thickness Predictor}
Optical spectra are simultaneously modulated by the macroscopic scalar film thickness and the continuous optical constants. Forcing a single network to resolve both modalities inherently exacerbates the one-to-many ambiguity. To address this issue, our framework introduces an independent thickness network denoted as $g_\phi$. Utilizing a Transformer architecture \cite{vaswani2017attention}, this branch extracts features from the incident ellipsometry parameters to produce a deterministic scalar prediction for the film thickness:
\begin{equation}
    \hat{\tilde{d}} = g_\phi(\tilde{\mathbf{x}}) \in \mathbb{R}^{B \times 1}
\end{equation}
By isolating the thickness prediction into a dedicated regression task, we substantially reduce the geometric uncertainty from the generative process. This predicted thickness subsequently serves as an absolutely critical and robust physical context $\tilde{d}_{ctx}$ to guide the generation of the remaining optical constants.

\subsubsection{Conditional Vector Field Network and Feature Modulation}
With the geometric ambiguity resolved, the generation of the optical constants $\tilde{\mathbf{y}} = [\tilde{n}, \tilde{k}]$ is handled by the flow matching network, which learns the velocity field $\hat{\mathbf{u}} = v_\theta(\tilde{\mathbf{y}}_t, t; \tilde{\mathbf{x}}, \tilde{d}_{ctx})$. To integrate these diverse modalities, we process the intermediate state and input features through distinct linear projections to form their computational tokens $\mathbf{tok}_y$ and $\mathbf{tok}_x$. Simultaneously, the continuous time variable $t$ is encoded into the time token $\mathbf{tok}_t$ via a sinusoidal time embedding followed by a linear projection. These tokens are then concatenated to construct the input sequence $\mathbf{S}_0 = [\mathbf{tok}_y, \mathbf{tok}_x, \mathbf{tok}_t]$. After adding learnable token type embeddings and applying layer normalization, the sequence is processed by a Transformer Encoder \cite{vaswani2017attention} to extract the readout representation $\mathbf{h}_0$.

Rather than naively concatenating the thickness condition, we employ a Feature-wise Linear Modulation (FiLM) mechanism \cite{perez2018film}. The thickness context $\tilde{d}_{ctx}$ is projected through a multilayer perceptron to scaling and shift parameters $\gamma$ and $\beta$. The readout feature is then modulated as:
\begin{equation}
    \text{FiLM}(\mathbf{h}_0, \tilde{d}_{ctx}) = (1 + \tanh(\gamma)) \odot \mathbf{h}_0 + \beta
\end{equation}
The final velocity prediction is obtained via a linear projection of this modulated feature. By modulating the readout features based on the thickness, the vector field is explicitly and guided toward the correct physical subspace.

\subsubsection{Joint Training with Gradient Detachment.}
A critical engineering challenge in our dual-branch architecture is training stability: if the flow-matching network backpropagates its error gradients through the thickness condition \(\tilde{d}_{\mathrm{ctx}}\), it can severely disrupt the convergence of the thickness predictor.
We therefore introduce a gradient detachment strategy using a \(\mathrm{stopgrad}(\cdot)\) operator.
During training, we sample a binary mask \(z \sim \mathrm{Bernoulli}(p_{\mathrm{TF}})\) following an annealed teacher-forcing schedule, and construct the conditioning context as
\begin{equation}
\tilde{d}_{\mathrm{ctx}} = z \cdot \tilde{d} + (1-z)\cdot \mathrm{stopgrad}(\hat{\tilde{d}}),
\label{eq:joint_detach}
\end{equation}
where \(\hat{\tilde{d}} = d_{\text{net}}(x)\) is the thickness predicted by the thickness network and \(\tilde{d}\) denotes the ground-truth thickness in the same scaled representation.
Furthermore, to prevent the flow-matching network from over-relying on the thickness condition, we inject sample-wise Gaussian noise \(\epsilon \sim \mathcal{N}(0,1)\) as \(\tilde{d}_{\mathrm{ctx}} \leftarrow \tilde{d}_{\mathrm{ctx}} + \sigma_d \epsilon\), and apply condition dropout at the batch level.
When condition dropout is triggered, we set \(\tilde{d}_{\mathrm{ctx}}=\varnothing\) and skip the FiLM, forcing the learned vector field to remain effective even without thickness cues.
At evaluation time, we disable teacher forcing and stochastic regularization (i.e., \(p_{\mathrm{TF}}=0\), no noise, and no dropout), so \(\tilde{d}_{\mathrm{ctx}}\) provides a deterministic conditioning signal.

\subsubsection{Optimization Objectives and Physics Constraints}
The entire framework is trained end-to-end through a comprehensive joint optimization strategy. First, the thickness network is supervised using a mean squared error loss against the true film thickness, denoted as $\mathcal{L}_{dnet} = \text{MSE}(\hat{\tilde{d}}, \tilde{d})$. Second, the flow matching network minimizes the velocity field error:
\begin{equation}
    \mathcal{L}_{FM} = \left\| \hat{\mathbf{u}} - \mathbf{u}^\star \right\|_2^2
\end{equation}

Finally, to guarantee that the generated optical parameters maintain absolute adherence to the laws of optics, we incorporate an optional physics reconstruction loss $\mathcal{L}_{recon}$. The predicted parameters are fed back into the forward Fresnel equations to reconstruct the complex reflection ratio. The total objective function is expressed as:
\begin{equation}
    \mathcal{L} = \mathcal{L}_{FM} + \lambda_d \mathcal{L}_{dnet} + \alpha_{recon} \mathcal{L}_{recon}
\end{equation}
This joint optimization ensures that the learned continuous mappings are statistically accurate while constrained by fundamental electromagnetic principles \cite{raissi2019physics}. The detailed mathematical formulations and computational procedures for all loss components are comprehensively documented in the Appendix.

\section{Experiments}
\label{experiments}

In this section, we systematically evaluate the performance of the proposed method on the EllipBench. We provide a comprehensive comparison with a wide range of baseline methods and conduct ablation studies to verify the effectiveness of each component.

\subsection{Experiment Settings}
\label{experiment settings}

To comprehensively assess our proposed framework, we configured a rigorous experimental setup. The EllipBench dataset was partitioned into training, validation, and testing sets using an $8:1:1$ ratio. We ensured a balanced distribution of all material categories across these subsets to prevent sampling bias. 

\subsubsection{Implementation Details}
Our framework was implemented using the PyTorch library and optimized via the Adam algorithm to ensure stable gradient estimation over the diverse material dataset. During the inference phase of the flow matching network, we utilized numerical integration to sample the continuous vector field and seamlessly construct the final optical parameters. Exhaustive implementation details are provided in the Appendix.

\subsubsection{Evaluation Metrics}
For the evaluation metrics, we quantified the regression performance for each of the target physical parameters independently. Specifically, to evaluate the predictive accuracy and variance explanation for the refractive index $n_2$, the extinction coefficient $k_2$, and the film thickness $d$, we employed three standard regression metrics for each variable. These metrics comprise the Mean Absolute Error (MAE), the Root Mean Square Error (RMSE), and the coefficient of determination $R^2$. The detailed mathematical formulations for these evaluation metrics are provided in the Appendix.

\begin{table}[t]
\centering
\caption{\textbf{Performance Comparison of Baseline Methods on EllipBench.} 
We report the comprehensive evaluation metrics across four progressive methodological paradigms. \textbf{Bold} indicates the optimal performance in each column.}
\label{tab:baselines_comparison}

\resizebox{\columnwidth}{!}{%
\begin{tabular}{l ccc ccc ccc}
\toprule

\multirow{2.5}{*}{\textbf{Method}} & 
\multicolumn{3}{c}{\textbf{Refractive Index ($n_2$)}} & 
\multicolumn{3}{c}{\textbf{Extinction Coeff. ($k_2$)}} & 
\multicolumn{3}{c}{\textbf{Thickness ($d$)}} \\
\cmidrule(lr){2-4} \cmidrule(lr){5-7} \cmidrule(lr){8-10}
& \textbf{MAE $\downarrow$} & \textbf{RMSE $\downarrow$} & \textbf{$R^2 \uparrow$} & \textbf{MAE $\downarrow$} & \textbf{RMSE $\downarrow$} & \textbf{$R^2 \uparrow$} & \textbf{MAE $\downarrow$} & \textbf{RMSE $\downarrow$} & \textbf{$R^2 \uparrow$} \\
\midrule

\rowcolor{gray!15} 
\multicolumn{10}{c}{\textbf{\textsc{Group I: Conventional Regression Method}}} \\
\addlinespace[0.2em]

Ridge Regression         & 0.927 & 1.260 & 0.137 & 1.288 & 1.661 & 0.362 & 2.097 & 2.520 & 0.237 \\
ElasticNet Regression    & 0.927 & 1.260 & 0.137 & 1.288 & 1.661 & 0.362 & 2.097 & 2.520 & 0.237 \\
Support Vector Regressor & 0.410 & 0.663 & 0.761 & 0.681 & 1.225 & 0.653 & 1.638 & 2.181 & 0.428 \\
\cmidrule(lr){1-10} 

\rowcolor{gray!15} 
\multicolumn{10}{c}{\textbf{\textsc{Group II: Ensemble Learning Method}}} \\
\addlinespace[0.2em]

Random Forest            & 0.246 & 0.441 & 0.894 & 0.394 & \textbf{0.748} & \textbf{0.871} & 0.955 & 1.444 & 0.749 \\
XGBoost                  & 0.319 & 0.495 & 0.867 & 0.532 & 0.873 & 0.824 & 1.287 & 1.713 & 0.648 \\
LightGBM                 & 0.313 & 0.489 & 0.870 & 0.524 & 0.861 & 0.829 & 1.264 & 1.685 & 0.659 \\
\cmidrule(lr){1-10}

\rowcolor{gray!15} 
\multicolumn{10}{c}{\textbf{\textsc{Group III: Deep Learning Method}}} \\
\addlinespace[0.2em]

MLP                      & 0.278 & 0.437 & 0.896 & 0.450 & 0.770 & 0.863 & 1.066 & 1.483 & 0.735 \\
Transformer              & 0.279 & 0.441 & 0.894 & 0.464 & 0.776 & 0.861 & 1.006 & 1.443 & 0.749 \\
\cmidrule(lr){1-10}

\rowcolor{gray!15} 
\multicolumn{10}{c}{\textbf{\textsc{Group IV: Physics Informed Method}}} \\
\addlinespace[0.2em]

PINN                     & 0.622 & 0.882 & 0.577 & 0.788 & 1.133 & 0.703 & 1.396 & 1.838 & 0.594 \\
\textbf{DCFM (Ours)}     & \textbf{0.237} & \textbf{0.428} & \textbf{0.900} & \textbf{0.390} & 0.764 & 0.865 & \textbf{0.807} & \textbf{1.192} & \textbf{0.829} \\

\bottomrule
\end{tabular}%
}
\end{table}
\subsubsection{Baseline Methods}
To benchmark our approach, we selected a diverse array of algorithms strategically categorized into four paradigms. These comprise conventional regression models (Ridge Regression and ElasticNet) for establishing linear mapping baselines, and robust tree-based ensemble learning methods (Random Forest, XGBoost, and LightGBM) tailored for numerical tabular data. For deep learning baselines, we evaluated purely data-driven structures, specifically Multi-Layer Perceptrons (MLP) \cite{lecun2015deep} and standard Transformer \cite{vaswani2017attention}, which rely on large-scale datasets to learn deterministic mappings. Finally, we implemented a Physics-Informed Neural Network (PINN) \cite{raissi2019physics} as our advanced baseline, which integrates fundamental electromagnetic equations into its optimization process through specialized penalty terms. Comprehensive implementation details, network architectures, and hyperparameter configurations for all evaluated baseline models are thoroughly documented in the Appendix.

\subsection{Main Results}
\label{main results}

Table~\ref{tab:baselines_comparison} reports the performance of all compared methods on EllipBench. DCFM achieves the strongest overall performance among the compared methods, attaining the lowest MAE and RMSE and the highest $R^2$ for refractive index $n_2$ and thickness $d$, while remaining competitive on extinction coefficient $k_2$. Conventional regression methods perform worst, showing that simple deterministic mappings are inadequate for the strong nonlinearity of inverse ellipsometry. Ensemble methods, including Random Forest, XGBoost, and LightGBM, improve substantially over linear baselines, but still remain limited, particularly for the joint recovery of optical constants and thickness. Among them, Random Forest is the strongest baseline, achieving competitive results on $n_2$ and $k_2$, yet still trailing DCFM, especially in thickness prediction.

Deep learning models further improve performance by leveraging large-scale data. Both MLP and Transformer outperform most conventional baselines on optical-constant prediction, but their gains on thickness estimation are modest, indicating that standard discriminative models still struggle with the ambiguity of the inverse mapping. PINN also fails to outperform purely data-driven deep models, despite incorporating explicit physical constraints. This suggests that physics regularization alone is insufficient to resolve the ill-posedness of inverse ellipsometry. In contrast, DCFM delivers the most robust and balanced performance, with particularly strong gains in thickness recovery: compared with the strongest deep baseline, it reduces the thickness MAE from 1.006 to 0.807 and improves $R^2$ from 0.749 to 0.829. These results show that explicitly modeling the inverse distribution under decoupled geometric conditioning is more effective than deterministic regression or constraint-only formulations.
\begin{table}[t]
\centering
\scriptsize
\renewcommand{\arraystretch}{1.25} 
\setlength{\tabcolsep}{3pt} 

\caption{\textbf{Ablation Study on Core Components of DCFM.} 
We report the evaluation metrics for different architectural variants. The notation w/o denotes the removal of a specific component from the full model. \textbf{Bold} indicates the optimal performance.}
\label{tab:ablation_study}

\resizebox{\columnwidth}{!}{%
\begin{tabular}{@{} l ccc ccc ccc @{}}
\toprule

\multirow{2.5}{*}{\textbf{Model Variant}} & 
\multicolumn{3}{c}{\textbf{Refractive Index ($n_2$)}} & 
\multicolumn{3}{c}{\textbf{Extinction Coeff. ($k_2$)}} & 
\multicolumn{3}{c}{\textbf{Thickness ($d$)}} \\
\cmidrule(lr){2-4} \cmidrule(lr){5-7} \cmidrule(lr){8-10}
& \textbf{MAE} $\downarrow$ & \textbf{RMSE} $\downarrow$ & \textbf{$R^2$} $\uparrow$ & \textbf{MAE} $\downarrow$ & \textbf{RMSE} $\downarrow$ & \textbf{$R^2$} $\uparrow$ & \textbf{MAE} $\downarrow$ & \textbf{RMSE} $\downarrow$ & \textbf{$R^2$} $\uparrow$ \\
\midrule

w/o Thickness Cond.      & 0.316 & 0.543 & 0.840 & 0.505 & 0.917 & 0.806 & \textbf{0.801} & 1.217 & 0.822 \\
w/o FiLM Modulation      & 0.270 & 0.474 & 0.878 & 0.402 & 0.784 & 0.858 & 0.841 & 1.264 & 0.808 \\
w/o Recon. Loss          & 0.251 & 0.432 & 0.898 & 0.392 & \textbf{0.760} & \textbf{0.866} & 0.806 & 1.222 & 0.820 \\
\cmidrule(lr){1-10} 
\textbf{Full DCFM (Ours)}& \textbf{0.237} & \textbf{0.428} & \textbf{0.900} & \textbf{0.390} & 0.764 & 0.865 & 0.807 & \textbf{1.192} & \textbf{0.829} \\

\bottomrule
\end{tabular}%
}
\end{table}


\subsection{Ablation Study}
\label{ablation study}

To validate our core architectural innovations, we evaluated three ablated variants of our framework, with details presented in Table \ref{tab:ablation_study}. First, removing the geometric thickness condition forced the flow matching network to rely purely on spectral inputs without any physical prior. This unconditioned variant suffered a severe degradation in optical constant predictions, notably increasing the MAE for the extinction coefficient from 0.390 to 0.505, while the independent thickness prediction remained stable. This result suggests that supplying a decoupled geometric scalar is important to resolve the inherent one-to-many mapping ambiguity.

Second, replacing the Feature-wise Linear Modulation with a naive token concatenation strategy caused comprehensive performance drops across all physical parameters, confirming that dynamically scaling hidden representations is significantly more effective than simple token aggregation. Finally, eliminating the Fresnel based physics reconstruction loss increased the overall prediction variance, specifically elevating the RMSE for film thickness and refractive index. This highlights its critical role as a physical regularizer that binds the continuous mappings to fundamental electromagnetic principles and ensures structural stability.


\section{Conclusion}
\label{conclusion}

In conclusion, this paper introduces \textbf{EllipBench}, a large-scale open-source benchmark covering diverse materials and substrate configurations, together with a novel physical metric based on energy conservation error. To address the one-to-many ambiguity inherent in inverse ellipsometry, we further propose the \textbf{Decoupled Conditional Flow Matching (DCFM)} framework, which explicitly decouples geometric thickness and injects it as a physical prior through Feature-wise Linear Modulation, thereby modeling the complex inverse distribution of optical constants and achieving state-of-the-art predictive performance on EllipBench. Notably, DCFM is designed for dense-spectrum inversion, where \((\Psi,\Delta)\) are observed on a predefined wavelength grid and \((n,k)\) are predicted in a point-wise manner conditioned on thickness. While this formulation aligns well with our benchmark and supports scalable training on millions of samples, it also defines the current scope of the method: scenarios involving sparse, irregular, or missing wavelength measurements require exploiting spectral coherence at the function level, which our present model does not explicitly enforce across wavelengths. Since EllipBench can naturally support settings in which only a subset of wavelength points is observed and the full spectral segment must be inferred, an important direction for future work is to extend DCFM toward sparse-to-dense spectral reconstruction using masked conditioning, sequence models, or operator-style parameterizations. We hope that EllipBench will provide a strong foundation for and stimulate further research on partial-measurement ellipsometry inversion.

\begin{credits}
\subsubsection{\ackname}
Omitted for double-blind review.

\subsubsection{\discintname}
The authors have no competing interests to declare that are relevant to the content of this article.
\end{credits}

\bibliographystyle{splncs04}
\bibliography{reference}

@article{ongie2020deep,
  title={Deep learning techniques for inverse problems in imaging},
  author={Ongie, Gregory and Jalal, Ajil and Metzler, Christopher A and Baraniuk, Richard G and Dimakis, Alexandros G and Willett, Rebecca},
  journal={IEEE Journal on Selected Areas in Information Theory},
  volume={1},
  number={1},
  pages={39--56},
  year={2020},
  publisher={IEEE}
}

@article{vaswani2017attention,
  title={Attention is all you need},
  author={Vaswani, Ashish and Shazeer, Noam and Parmar, Niki and Uszkoreit, Jakob and Jones, Llion and Gomez, Aidan N and Kaiser, {\L}ukasz and Polosukhin, Illia},
  journal={Advances in neural information processing systems},
  volume={30},
  year={2017}
}

@inproceedings{perez2018film,
  title={FiLM: Visual Reasoning with a General Conditioning Layer},
  author={Perez, Ethan and Strub, Florian and de Vries, Harm and Dumoulin, Vincent and Courville, Aaron},
  booktitle={Proceedings of the AAAI Conference on Artificial Intelligence},
  volume={32},
  number={1},
  year={2018}
}

@book{fujiwara2007spectroscopic,
  title={Spectroscopic ellipsometry: principles and applications},
  author={Fujiwara, Hiroyuki},
  year={2007},
  publisher={John Wiley \& Sons}
}

@inproceedings{albergo2023building,
  title={Building Normalizing Flows with Stochastic Interpolants},
  author={Albergo, Michael and Vanden-Eijnden, Eric},
  booktitle={ICLR 2023 Conference},
  year={2023}
}

@inproceedings{lipmanflow,
  title={Flow Matching for Generative Modeling},
  author={Lipman, Yaron and Chen, Ricky TQ and Ben-Hamu, Heli and Nickel, Maximilian and Le, Matthew},
  booktitle={The Eleventh International Conference on Learning Representations}
}

@article{liu2018training,
  title={Training deep neural networks for the inverse design of nanophotonic structures},
  author={Liu, Dianjing and Tan, Yixuan and Khoram, Erfan and Yu, Zongfu},
  journal={Acs Photonics},
  volume={5},
  number={4},
  pages={1365--1369},
  year={2018},
  publisher={ACS Publications}
}

@article{carleo2019machine,
  title={Machine learning and the physical sciences},
  author={Carleo, Giuseppe and Cirac, Ignacio and Cranmer, Kyle and Daudet, Laurent and Schuld, Maria and Tishby, Naftali and Vogt-Maranto, Leslie and Zdeborov{\'a}, Lenka},
  journal={Reviews of Modern Physics},
  volume={91},
  number={4},
  pages={045002},
  year={2019},
  publisher={APS}
}

@article{jellison1993data,
  title={Data analysis for spectroscopic ellipsometry},
  author={Jellison Jr, Gerald E},
  journal={Thin Solid Films},
  volume={234},
  number={1-2},
  pages={416--422},
  year={1993},
  publisher={Elsevier}
}

@article{jellison1998spectroscopic,
  title={Spectroscopic ellipsometry data analysis: measured versus calculated quantities},
  author={Jellison Jr, Gerald E},
  journal={Thin solid films},
  volume={313},
  pages={33--39},
  year={1998},
  publisher={Elsevier}
}

@article{merchant2023scaling,
  title={Scaling deep learning for materials discovery},
  author={Merchant, Amil and Batzner, Simon and Schoenholz, Samuel S and Aykol, Muratahan and Cheon, Gowoon and Cubuk, Ekin Dogus},
  journal={Nature},
  volume={624},
  number={7990},
  pages={80--85},
  year={2023},
  publisher={Nature Publishing Group UK London}
}

@article{jumper2021highly,
  title={Highly accurate protein structure prediction with AlphaFold},
  author={Jumper, John and Evans, Richard and Pritzel, Alexander and Green, Tim and Figurnov, Michael and Ronneberger, Olaf and Tunyasuvunakool, Kathryn and Bates, Russ and {\v{Z}}{\'\i}dek, Augustin and Potapenko, Anna and others},
  journal={nature},
  volume={596},
  number={7873},
  pages={583--589},
  year={2021},
  publisher={Nature Publishing Group UK London}
}

@article{stokes2020deep,
  title={A deep learning approach to antibiotic discovery},
  author={Stokes, Jonathan M and Yang, Kevin and Swanson, Kyle and Jin, Wengong and Cubillos-Ruiz, Andres and Donghia, Nina M and MacNair, Craig R and French, Shawn and Carfrae, Lindsey A and Bloom-Ackermann, Zohar and others},
  journal={Cell},
  volume={180},
  number={4},
  pages={688--702},
  year={2020},
  publisher={Elsevier}
}

@ArtifactSoftware{R,
    title = {R: A Language and Environment for Statistical Computing},
    author = {{R Core Team}},
    organization = {R Foundation for Statistical Computing},
    address = {Vienna, Austria},
    year = {2019},
    url = {https://www.R-project.org/},
}

@article{liu2021machine,
  title={Machine learning powered ellipsometry},
  author={Liu, Jinchao and Zhang, Di and Yu, Dianqiang and Ren, Mengxin and Xu, Jingjun},
  journal={Light: Science \& Applications},
  volume={10},
  number={1},
  pages={55},
  year={2021},
  publisher={Nature Publishing Group UK London}
}

@misc{bell1978solutions,
  title={Solutions of Ill-Posed Problems.},
  author={Bell, John B},
  year={1978},
  publisher={JSTOR}
}

@inproceedings{akbalik2009inverse,
  title={An inverse ellipsometric problem for thin film characterization: comparison of different optimization methods},
  author={Akbal{\i}k, Ay{\c{s}}e and Soulan, S{\'e}bastien and Tortai, Jean-Herv{\'e} and Fuard, David and Kone, Issiaka and Hazart, J{\'e}r{\^o}me and Schiavone, Patrick},
  booktitle={Metrology, Inspection, and Process Control for Microlithography XXIII},
  volume={7272},
  pages={1122--1128},
  year={2009},
  organization={SPIE},
  publisher={}
}

@article{lecun2015deep,
  title={Deep learning},
  author={LeCun, Yann and Bengio, Yoshua and Hinton, Geoffrey},
  journal={nature},
  volume={521},
  number={7553},
  pages={436--444},
  year={2015},
  publisher={Nature Publishing Group UK London}
}

@article{urban1994real,
  title={Real time, in-situ ellipsometry solutions using artificial neural network pre-processing},
  author={Urban III, Frank K and Tabet, Milad F},
  journal={Thin Solid Films},
  volume={245},
  number={1-2},
  pages={167--173},
  year={1994},
  publisher={Elsevier}
}

@article{collins2000recent,
  title={Recent progress in thin film growth analysis by multichannel spectroscopic ellipsometry},
  author={Collins, RW and Koh, Joohyun and Fujiwara, H and Rovira, PI and Ferlauto, AS and Zapien, JA and Wronski, CR and Messier, R},
  journal={Applied surface science},
  volume={154},
  pages={217--228},
  year={2000},
  publisher={Elsevier}
}

@article{zhu2024lightweight,
  title={A Lightweight Neural Network for Spectroscopic Ellipsometry Analysis},
  author={Zhu, Pengfei and Zhang, Di and Niu, Xin and Liu, Jinchao and Ren, Mengxin and Xu, Jingjun},
  journal={Advanced Optical Materials},
  pages={2301381},
  year={2024},
  publisher={Wiley Online Library}
}

@article{oates2011characterization,
  title={Characterization of plasmonic effects in thin films and metamaterials using spectroscopic ellipsometry},
  author={Oates, TWH and Wormeester, Herbert and Arwin, Hans},
  journal={Progress in Surface Science},
  volume={86},
  number={11-12},
  pages={328--376},
  year={2011},
  publisher={Elsevier}
}

@article{wei2019machine,
  title={Machine learning in materials science},
  author={Wei, Jing and Chu, Xuan and Sun, Xiang-Yu and Xu, Kun and Deng, Hui-Xiong and Chen, Jigen and Wei, Zhongming and Lei, Ming},
  journal={InfoMat},
  volume={1},
  number={3},
  pages={338--358},
  year={2019},
  publisher={Wiley Online Library}
}

@article{raissi2019physics,
  title={Physics-informed neural networks: A deep learning framework for solving forward and inverse problems involving nonlinear partial differential equations},
  author={Raissi, Maziar and Perdikaris, Paris and Karniadakis, George E},
  journal={Journal of Computational physics},
  volume={378},
  pages={686--707},
  year={2019},
  publisher={Elsevier}
}

@article{wang2023ellipsonet,
  title={EllipsoNet: Deep-learning-enabled optical ellipsometry for complex thin films},
  author={Wang, Ziyang},
  journal={Bulletin of the American Physical Society},
  volume={68},
  year={2023},
  publisher={APS}
}

@article{adler2017solving,
  title={Solving ill-posed inverse problems using iterative deep neural networks},
  author={Adler, Jonas and {\"O}ktem, Ozan},
  journal={Inverse Problems},
  volume={33},
  number={12},
  pages={124007},
  year={2017},
  publisher={IOP Publishing}
}

@article{johs1993regression,
  title={Regression calibration method for rotating element ellipsometers},
  author={Johs, Blaine},
  journal={Thin Solid Films},
  volume={234},
  number={1-2},
  pages={395--398},
  year={1993},
  publisher={Elsevier}
}

@article{poelman2003methods,
  title={Methods for the determination of the optical constants of thin films from single transmission measurements: a critical review},
  author={Poelman, Dirk and Smet, Philippe Frederic},
  journal={Journal of Physics D: Applied Physics},
  volume={36},
  number={15},
  pages={1850--1857},
  year={2003}
}

@book{tompkins2005handbook,
  title={Handbook of ellipsometry},
  author={Tompkins, Harland and Irene, Eugene A},
  year={2005},
  publisher={William Andrew}
}

@book{losurdo2013ellipsometry,
  title={Ellipsometry at the Nanoscale},
  author={Losurdo, Maria and Hingerl, Kurt},
  volume={268},
  year={2013},
  publisher={Springer}
}

@book{macleod2010thin,
  title={Thin-film optical filters},
  author={Macleod, H Angus and Macleod, H Angus},
  year={2010},
  publisher={CRC press}
}

@misc{azzam1978ellipsometry,
  title={Ellipsometry and polarized light},
  author={Azzam, Rasheed MA and Bashara, Nicholas Mitchell and Ballard, Stanley S},
  year={1978},
  publisher={American Institute of Physics}
}

@article{chen2018neural,
  title={Neural ordinary differential equations},
  author={Chen, Ricky TQ and Rubanova, Yulia and Bettencourt, Jesse and Duvenaud, David K},
  journal={Advances in neural information processing systems},
  volume={31},
  year={2018}
}

@article{arunachalam2022machine,
  title={Machine learning-enhanced efficient spectroscopic ellipsometry modeling},
  author={Arunachalam, Ayush and Berriel, S Novia and Banerjee, Parag and Basu, Kanad},
  journal={arXiv preprint arXiv:2201.04933},
  year={2022}
}

@article{jiang2024generic,
  title={Generic characterization method for nano-gratings using deep-neural-network-assisted ellipsometry},
  author={Jiang, Zijie and Gan, Zhuofei and Liang, Chuwei and Li, Wen-Di},
  journal={Nanophotonics},
  number={0},
  year={2024},
  publisher={De Gruyter}
}

@book{lucarini2005kramers,
  title={Kramers-Kronig relations in optical materials research},
  author={Lucarini, Valerio and Peiponen, Kai-Erik and Saarinen, Jarkko J and Vartiainen, Erik M},
  publisher={Springer}
}

@article{wang2023measuring,
  title={Measuring complex refractive index through deep-learning-enabled optical reflectometry},
  author={Wang, Ziyang and Lin, Yuxuan Cosmi and Zhang, Kunyan and Wu, Wenjing and Huang, Shengxi},
  journal={2D Materials},
  volume={10},
  number={2},
  pages={025025},
  year={2023},
  publisher={IOP Publishing}
}

@article{karniadakis2021physics,
  title={Physics-informed machine learning},
  author={Karniadakis, George Em and Kevrekidis, Ioannis G and Lu, Lu and Perdikaris, Paris and Wang, Sifan and Yang, Liu},
  journal={Nature Reviews Physics},
  volume={3},
  number={6},
  pages={422--440},
  year={2021},
  publisher={Nature Publishing Group UK London}
}

@article{ardizzone2018analyzing,
  title={Analyzing inverse problems with invertible neural networks},
  author={Ardizzone, Lynton and Kruse, Jakob and Wirkert, Sebastian and Rahner, Daniel and Pellegrini, Eric W and Klessen, Ralf S and Maier-Hein, Lena and Rother, Carsten and K{\"o}the, Ullrich},
  journal={arXiv preprint arXiv:1808.04730},
  year={2018}
}

@article{kobyzev2020normalizing,
  title={Normalizing flows: An introduction and review of current methods},
  author={Kobyzev, Ivan and Prince, Simon JD and Brubaker, Marcus A},
  journal={IEEE transactions on pattern analysis and machine intelligence},
  volume={43},
  number={11},
  pages={3964--3979},
  year={2020},
  publisher={IEEE}
}

\end{document}


\clearpage
\appendix

\begin{center}
    \LARGE \textbf{Appendix for \\ Modeling Inverse Ellipsometry Problem via Flow Matching with a Large-Scale Dataset}
\end{center}
\vspace{1em}

\section{Forward Ellipsometry Model Derivations}
\label{app:forward_model}
The exact forward mapping from the physical properties of thin films to the measurable ellipsometric parameters $\Psi$ and $\Delta$ relies on electromagnetic wave propagation governed by Maxwell equations. We denote the refractive index and extinction coefficient of the thin film as $n_2$ and $k_2$, and the corresponding optical constants of the substrate as $n_3$ and $k_3$. The complex refractive indices for the ambient air environment, the thin film, and the substrate are defined as $N_1$, $N_2$, and $N_3$ respectively. The light propagation angles $\theta_1$, $\theta_2$, and $\theta_3$ across these interfaces are governed by Snell law of refraction:
\begin{equation}
\begin{split}
    N_2 &= n_2 + i k_2, \\ 
    N_3 &= n_3 + i k_3, \\
    N_1 \sin \theta_1 &= N_2 \sin \theta_2 = N_3 \sin \theta_3
\end{split}
\end{equation}
When evaluating the complex angles of refraction, the root selection strictly enforces a positive imaginary component for the cosine terms to guarantee correct physical wave attenuation and continuous forward propagation direction.

Subsequently, the single layer reflection coefficients ($r_p, r_s$) and transmission coefficients ($t_p, t_s$) for parallel and perpendicularly polarized light at a given interface are computed via the Fresnel equations. Using the subscripts up and down to indicate the relative positions of the adjacent materials in the stratified model, these coefficients are formulated as:
\begin{equation}
\begin{split}
    r_p = & \frac{N_{\text{down}} \cos \theta_{\text{up}} - N_{\text{up}} \cos \theta_{\text{down}}}{N_{\text{down}} \cos \theta_{\text{up}} + N_{\text{up}} \cos \theta_{\text{down}}}, \quad
    t_p = \frac{2 N_{\text{up}} \cos \theta_{\text{up}}}{N_{\text{down}} \cos \theta_{\text{up}} + N_{\text{up}} \cos \theta_{\text{down}}} \\
    r_s = & \frac{N_{\text{up}} \cos \theta_{\text{up}} - N_{\text{down}} \cos \theta_{\text{down}}}{N_{\text{up}} \cos \theta_{\text{up}} + N_{\text{down}} \cos \theta_{\text{down}}}, \quad
    t_s = \frac{2 N_{\text{up}} \cos \theta_{\text{up}}}{N_{\text{up}} \cos \theta_{\text{up}} + N_{\text{down}} \cos \theta_{\text{down}}}
\end{split}
\end{equation}

For a double layer structural model comprising the ambient environment, the thin film, and the substrate, the total complex amplitude coefficients incorporate the infinite series of multiple internal reflections within the film layer. By adopting the positive time harmonic convention $e^{+i\omega t}$, the spatial propagation phase term accumulates as $e^{-ikz}$, dictating the negative sign in the exponential. The total reflection coefficients $r_{pp}$ and $r_{ss}$, alongside the total transmission coefficients $t_{pp}$ and $t_{ss}$, are expressed as:
\begin{equation}
\begin{split}
    r_{pp} &= \frac{r_p(1,2) + r_p(2,3) \exp(-i 2 \beta)}{1 + r_p(1,2) r_p(2,3) \exp(-i 2 \beta)}, \quad
    t_{pp} = \frac{t_p(1,2) t_p(2,3) \exp(-i \beta)}{1 + r_p(1,2) r_p(2,3) \exp(-i 2 \beta)} \\
    r_{ss} &= \frac{r_s(1,2) + r_s(2,3) \exp(-i 2 \beta)}{1 + r_s(1,2) r_s(2,3) \exp(-i 2 \beta)}, \quad
    t_{ss} = \frac{t_s(1,2) t_s(2,3) \exp(-i \beta)}{1 + r_s(1,2) r_s(2,3) \exp(-i 2 \beta)}
\end{split}
\label{eq:app_multilayer_formulas}
\end{equation}
where $\beta$ indicates the phase thickness governing the interference phenomena, calculated using the geometric film thickness $d$ and the incident wavelength $\lambda$:
\begin{equation}
    \beta = 2 \pi \left(\frac{d}{\lambda}\right) N_2 \cos \theta_2
\end{equation}
Finally, the standard ellipsometric measurement parameters $\Psi$ and $\Delta$ are intrinsically linked to the complex reflection ratio $\rho$ through the fundamental ellipsometry equation:
\begin{equation}
    \rho = \tan \Psi \exp(i \Delta) = \frac{r_{pp}}{r_{ss}}
\end{equation}

\section{Energy Conservation Error Formulation}
\label{app:ec_error}
To evaluate the physical consistency and intrinsic absorption characteristics of the synthesized dataset, we established the Energy Conservation Error. For ideal non absorbing ambient and substrate media ($k_1 = 0$ and $k_3 = 0$), the incident electromagnetic energy is strictly partitioned such that reflectance and transmittance sum to unity. However, for metallic and semiconductor films where the extinction coefficient $k_2 > 0$, inherent absorption $A \ge 0$ occurs, satisfying the generalized balance $R + T + A = 1$. 

Consequently, evaluating the deviation magnitude $|R + T - 1|$ acts as a error that quantifies this non absorption deviation. This formulation physically explains the macroscopically large statistical values observed in specific metallic combinations, reflecting substantial absorption and complex substrate coupling rather than strictly numerical errors. The fundamental Energy Conservation Error is evaluated across the dataset on a point-wise basis:
\begin{equation}
\text{EC Error} = \frac{1}{N_{total}}\sum_{s=1}^{N_{total}}\left(|R_{p,s} + T_{p,s} - 1| + |R_{s,s} + T_{s,s} - 1|\right)
\end{equation}
where $N_{total}$ denotes the total number of point-wise wavelength samples across all configurations. The overall reflectance is calculated directly from the magnitude squared of the complex reflection coefficients $R_p = |r_{pp}|^2$ and $R_s = |r_{ss}|^2$. 

Strictly speaking, absolute energy conservation is precisely defined solely for non absorbing ambient and substrate conditions. However, to maintain a consistent proxy metric across all structural configurations including absorbing substrates, the computation of the transmittance $T_p$ and $T_s$ rigorously isolates the real component of the complex refractive indices and complex propagation angles to accurately represent the active power flow matching across boundaries:
\begin{equation}
    T_p = \frac{\text{Re}(N_3\cos\theta_3)}{\text{Re}(N_1\cos\theta_1)}|t_{pp}|^2, \quad T_s = \frac{\text{Re}(N_3\cos\theta_3)}{\text{Re}(N_1\cos\theta_1)}|t_{ss}|^2
\end{equation}
During the dataset generation process, we utilized this strict definition to filter severe numerical ill posedness. Any simulated point-wise sample exhibiting an absolute, unscaled deviation magnitude greater than 0.001 was permanently discarded. For readability and comparative analysis, all EC Error reported in subsequent macroscopic histograms and heatmaps are scaled by a factor of $10^3$ relative to this foundational definition.

\section{Loss Functions and Optimization Objectives}
\label{app:losses}
The Decoupled Conditional Flow Matching framework is optimized using a unified gradient descent strategy combining multiple specialized loss components.

\textit{Thickness Regression Loss.} The decoupled geometric branch is trained using a standard mean squared error against the ground truth thickness distribution:
\begin{equation}
    \mathcal{L}_{dnet} = \frac{1}{B} \sum_{b=1}^{B} (\hat{d}^{(b)} - d^{(b)})^2
\end{equation}

\textit{Flow Matching Vector Field Loss.} The generative branch is optimized by minimizing the discrepancy between the network predicted vector field $\hat{\mathbf{u}}$ and the target linear interpolation velocity $\mathbf{u}^\star = \mathbf{y} - \epsilon$:
\begin{equation}
    \mathcal{L}_{FM} = \frac{1}{B} \sum_{b=1}^{B} \left\| v_\theta(\mathbf{y}_t^{(b)}, t^{(b)}; \mathbf{x}^{(b)}, d_{ctx}^{(b)}) - (\mathbf{y}^{(b)} - \epsilon^{(b)}) \right\|_2^2
\end{equation}

\textit{Physics Reconstruction Loss.} To enforce electromagnetic fidelity, we reproject the predicted physical properties $\hat{n}_2, \hat{k}_2, \hat{d}$ back to the complex reflection ratio space using the forward Fresnel formalism $F$. Assuming the measured ellipsometric angles $\Psi$ and $\Delta$ are provided in radians, the ground truth complex reflection ratio is rigorously constructed via Euler formula as $\rho_{true} = \tan\Psi(\cos\Delta + i\sin\Delta)$. To explicitly bypass severe discontinuities and phase unwrapping issues associated with the raw angular parameter $\Delta$, the complex mean squared error is computed independently over the real and imaginary components of the reflection ratio $\rho$:
\begin{equation}
    \mathcal{L}_{recon} = \frac{1}{B} \sum_{b=1}^{B} \left\| F_{Re,Im}(\hat{n}_2^{(b)}, \hat{k}_2^{(b)}, \hat{d}^{(b)}, n_3^{(b)}, k_3^{(b)}, \lambda^{(b)}) - \rho_{true, Re,Im}^{(b)} \right\|_2^2
\end{equation}
The total joint objective is a weighted summation of these components $\mathcal{L} = \mathcal{L}_{FM} + \lambda_d \mathcal{L}_{dnet} + \alpha_{recon} \mathcal{L}_{recon}$, where hyperparameter balancing is strictly maintained during training.

\section{Dataset Generation Details and Extended Analysis}
\label{app:dataset}
\textit{Generation Protocol.} EllipBench integrates diverse material systems by curating experimentally validated optical dispersion profiles for 98 thin film materials deposited on five prevalent substrates: amorphous Silicon, Indium Tin Oxide, Strontium Titanate, crystalline Silicon, and Polyimide. The spectral data for each material were sampled on a dense grid from 380.28 nm to 999.87 nm with a 2.6 nm resolution. We systematically varied the film thickness across 20 discrete intervals from 1 nm to 96 nm utilizing a precise logarithmic step. This progression is mathematically governed by the following formulation for interval index $j$ from 0 to 19:
\begin{equation}
    d_j = \exp\left(\log d_{\min} + j \frac{\log d_{\max} - \log d_{\min}}{19}\right)
\end{equation}
The forward simulation was executed at a fixed 70 degree incidence angle. Each material combination is divided in an 8 to 1 to 1 ratio into training, validation, and test sets. 

The stated 8 million entries strictly refer to individual point-wise spectral samples rather than aggregated full spectrum curves. Since the filtering protocol operates on a point-wise basis prior to dataset splitting, certain highly anomalous spectral points are excised. For sequence based modeling and continuous curve visualization tasks, these missing discrete wavelengths are reconstructed via a localized cubic polynomial interpolation. To explicitly avoid severe angular phase unwrapping discontinuities associated with the raw ellipsometric parameter $\Delta$, this polynomial interpolation is strictly applied independently to the real and imaginary components of the complex reflection ratio $\rho$, utilizing a symmetric sliding window encompassing four adjacent valid spectral points.

\begin{figure}[t]
\begin{center}
\centerline{\includegraphics[width=\columnwidth]{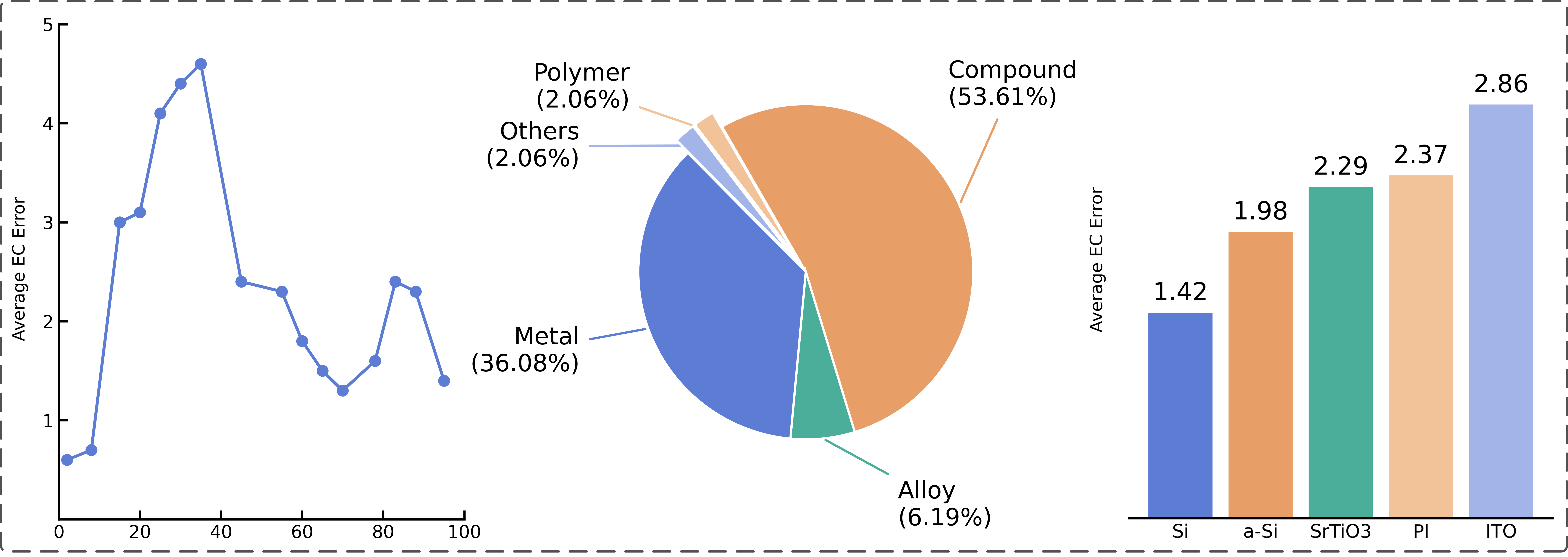}}
\caption{Comprehensive analysis of the optical thin film dataset. The left panel displays the non-linear correlation between Energy Conservation Error and film thickness. The middle panel illustrates the distribution of material categories within the dataset. The right panel compares the average Energy Conservation Error across different substrate configurations. Note that all EC Error values in this figure have been scaled by a factor of $10^3$ for readability.}
\label{fig:ec_error_overview}
\end{center}
\end{figure}

\textit{Extended Optical Property Analysis.} Figure \ref{fig:ec_error_overview} illustrates the optical property distribution of the 98 thin film materials in our dataset, spanning from dielectrics to semiconductors and metals. Refractive indices range from 1.2 to 4.5, with extinction coefficients from near zero to 5.0, ensuring comprehensive coverage of practical material properties. 

To provide a macroscopic overview of the dataset characteristics and physical consistency, Figure \ref{fig:ec_error_overview} visualizes the aggregate statistical metrics. As depicted in the middle panel, the dataset is predominantly composed of Compound (53.61 percent) and Metal (36.08 percent) categories, ensuring a robust representation of industrially relevant materials. The right panel demonstrates that substrate materials exert a significant influence on the overall measurement fidelity. Crystalline Silicon substrates exhibit superior performance with the lowest average scaled EC Error of 1.42, whereas Indium Tin Oxide substrates manifest the highest deviation magnitude at 2.86. Furthermore, the left panel reveals a non-linear correlation between the EC Error and film thickness, reaching a pronounced maximum in the 20 to 30 nm thickness region while displaying enhanced reliability in extremely thin or thicker domains. 

\begin{figure}[t]
  \centering
  \begin{minipage}{0.48\textwidth}
    \centering
    \includegraphics[width=\linewidth]{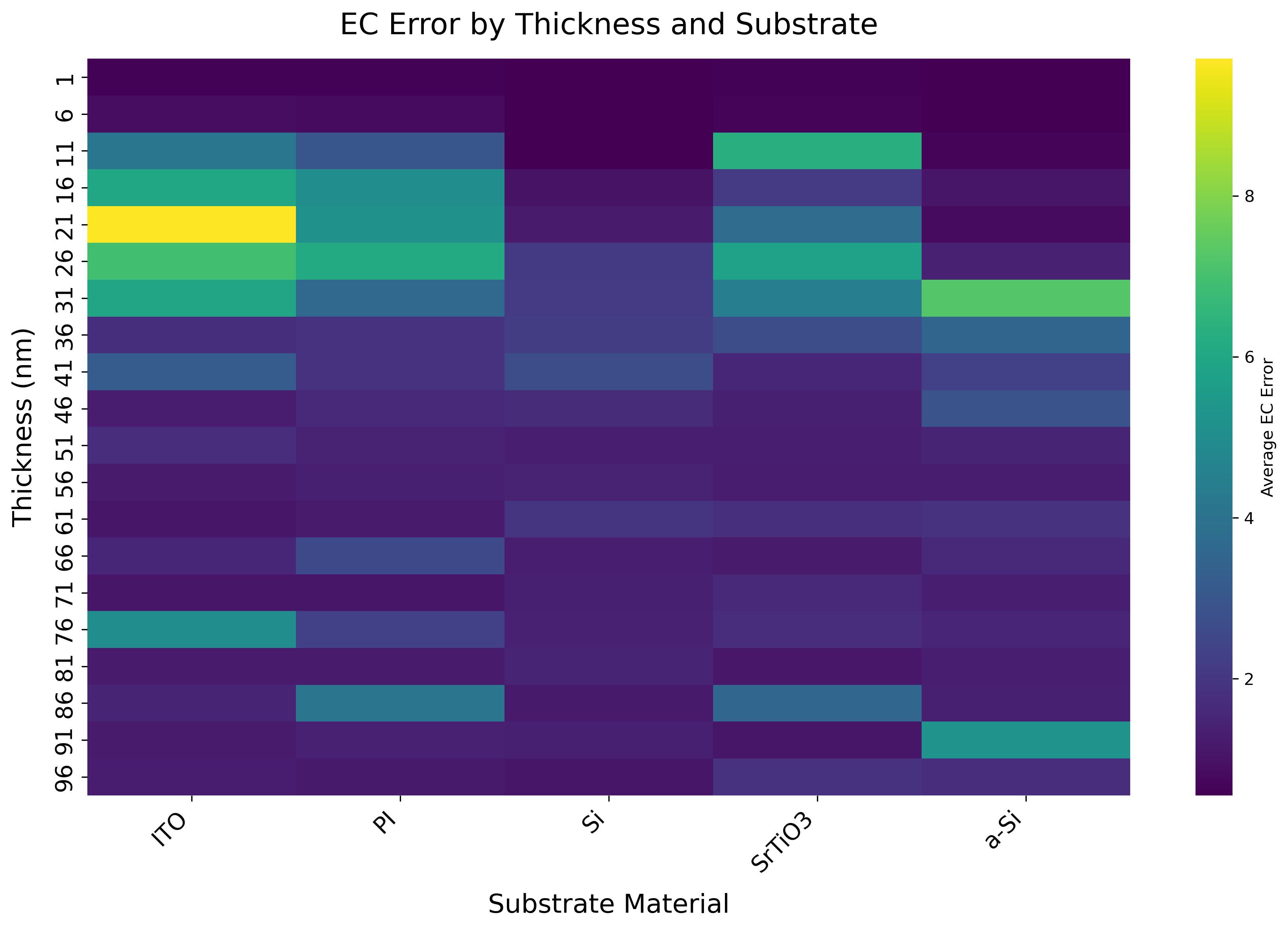}
    \caption{Energy Conservation Error Distribution Across Film Thickness and Substrate Materials (Scaled by $10^3$)}
    \label{fig:ec_error_thickness_substrate_app}
  \end{minipage}\hfill
  \begin{minipage}{0.48\textwidth}
    \centering
    \includegraphics[width=\linewidth]{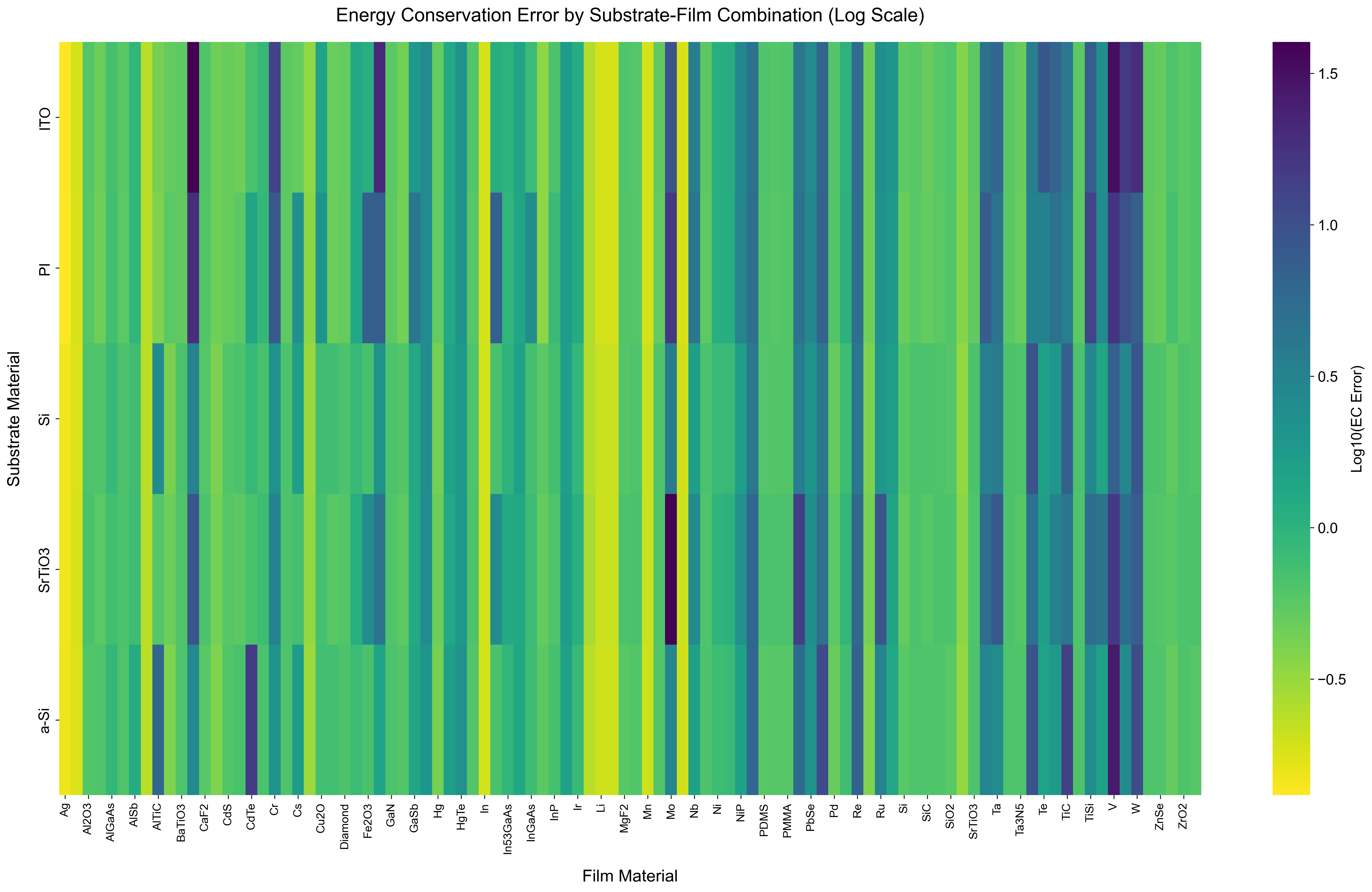}
    \caption{Logarithmic Visualization of Energy Conservation Error by Substrate Film Material Combinations (Scaled by $10^3$)}
    \label{fig:ec_error_substrate_film_app}
  \end{minipage}
\end{figure}

To further investigate these macroscopic phenomena, Figures \ref{fig:ec_error_thickness_substrate_app} and \ref{fig:ec_error_substrate_film_app} reveal specific measurement limitations through detailed EC Error visualization. Figure \ref{fig:ec_error_thickness_substrate_app} shows significant measurement deviations within the 16 to 31 nm thickness range, particularly on Indium Tin Oxide substrates. Films with 21 to 26 nm thickness consistently show elevated values across multiple substrates, pointing to systematic light matter interaction complexities rather than isolated experimental errors. Furthermore, the logarithmic visualization in Figure \ref{fig:ec_error_substrate_film_app} reveals that certain specific materials, namely Ag, Ir, and Mo, consistently exhibit high EC Error value across different substrates. Indium Tin Oxide substrates show higher measurement uncertainties with metallic films, highlighting extreme challenges in accurately characterizing metal transparent conductive oxide interfaces.

\section{Decoupled Conditional Flow Matching Implementation}
\label{app:dcfm_implementation}
The entire deep learning architecture was implemented utilizing the PyTorch framework.

\textit{Architecture Specifics.} The geometric thickness predictor employs a point-wise Transformer architecture with an embedding dimension of 128, 4 attention heads, and 4 sequential encoder layers with a feedforward dimension of 256. The optical constant flow matching network shares a parallel structural capacity. The input point-wise spectral features are mapped to distinct tokens and concatenated with a sinusoidal time embedding utilizing a dimension of 128. To inject the macroscopic thickness condition into the flow matching readout representation, the Feature wise Linear Modulation mechanism projects the scalar thickness into a 256 dimensional hidden space before regressing the affine scaling and shifting matrices.

\textit{Training Hyperparameters.} Network optimization was executed using the AdamW optimizer with a base learning rate of 1e-3 and a weight decay coefficient of 1e-4. We employed a batch size of 32 for continuous optimization over 50 epochs. A plateau based learning rate scheduler was integrated, reducing the learning rate by a factor of 0.5 upon validation metric stagnation for 10 consecutive epochs.

\textit{Stability Mechanisms.} To isolate the thickness branch and maintain gradient stability during joint training, we applied a probabilistic gradient detachment strategy formulated as:
\begin{equation}
    d_{ctx} = z \cdot d_{true} + (1 - z) \cdot \text{stopgrad}(\hat{d})
\end{equation}
where $z$ represents a Bernoulli mask governed by a progressive teacher forcing probability schedule that linearly decays from 1.0 to 0.0 across the training lifecycle. Furthermore, to prevent the flow matching network from over relying on the deterministic thickness condition, we injected a batch level condition dropout probability of 0.1, occasionally bypassing the modulation module entirely. When utilized, the continuous condition representation was dynamically perturbed using thickness condition noise sampled from a Gaussian distribution with a standard deviation of 0.01.

\textit{Inference Dynamics.} During the inference phase, the ordinary differential equation path is integrated from pure Gaussian noise using the Euler numerical method across 40 uniform time steps. The prediction variance is marginalized by drawing 4 independent sample trajectories per target instance and averaging the terminal state estimations.

\section{Baseline Architectures and Physics Informed Neural Network Details}
\label{app:baselines}
All conventional and tree based ensemble baselines were implemented using the Scikit Learn ecosystem. The continuous input physical features were rigorously standardized to a zero mean and unit variance normal distribution prior to regression.

\textit{Conventional and Ensemble Models.} The Ridge Regressor utilized an L2 regularization alpha of 1.0. The ElasticNet implementation balanced L1 and L2 penalties with a ratio of 0.5 and an alpha of 1e-3. The Support Vector Regressor employed a Radial Basis Function kernel configured with a penalty parameter C of 10.0 and an epsilon tube margin of 0.01. The Random Forest regressor aggregated 400 deep decision estimators. The XGBoost framework deployed 600 boosting rounds with a learning rate of 0.05, while the LightGBM model integrated 2000 trees utilizing a 0.03 learning rate.

\textit{Deep Learning Baselines.} The Multi Layer Perceptron baseline utilizes a point-wise architecture comprising 4 hidden layers with 256 dimensions each, Gaussian Error Linear Unit activations, and a dropout rate of 0.1. Similarly, the standard Transformer baseline processes point-wise spectral inputs through a distinct tokenization scheme, utilizing 4 attention heads and 4 encoder layers with a 128 dimensional embedding space and 256 dimensional feedforward networks. Unless otherwise specified, these deep learning baselines were trained utilizing the identical AdamW optimizer, batch size, and plateau based learning rate scheduling scheme as the DCFM framework.

\textit{Physics Informed Neural Network.} The Physics Informed Neural Network model is a specialized architecture designed to map the five dimensional input features to the three dimensional target parameters while incorporating strict physical law constraints. The network utilizes hyperbolic tangent activation functions to ensure smooth gradients required for calculating wave phenomena partial derivatives. After feature extraction, information flows through two specialized processing paths defined as the Optical Path and the Ellipsometry Path.

\begin{figure}[t]
\begin{center}
\centerline{\includegraphics[width=0.95\columnwidth]{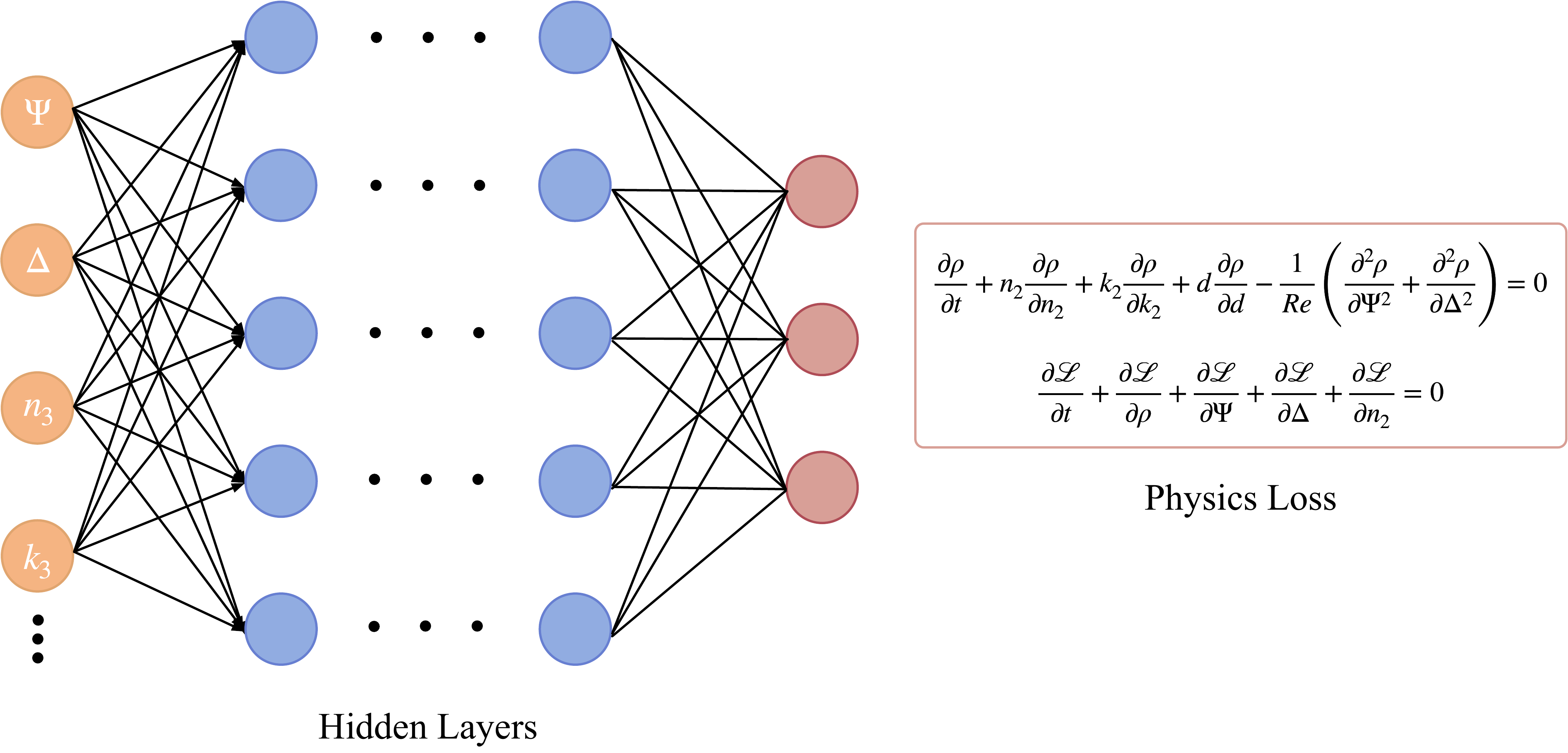}}
\caption{Physics Informed Neural Network Framework Overview}
\label{fig:pinn_framework}
\end{center}
\end{figure}

The profound innovation of this baseline lies in the partial differential equation constraints based on ellipsometry theory. The network explicitly models the spatial and temporal gradients of the complex reflection ratio:
\begin{equation}
\frac{\partial\rho}{\partial t} + n_2\frac{\partial\rho}{\partial n_2} + k_2\frac{\partial\rho}{\partial k_2} + d\frac{\partial\rho}{\partial d} - \frac{1}{Re}\left(\frac{\partial^2\rho}{\partial\Psi^2} + \frac{\partial^2\rho}{\partial\Delta^2}\right) = 0
\end{equation}
as well as a Lagrangian formulation representing the fundamental energy conservation embedded within the optical system:
\begin{equation}
\frac{\partial\mathcal{L}}{\partial t} + \frac{\partial\mathcal{L}}{\partial\rho} + \frac{\partial\mathcal{L}}{\partial\Psi} + \frac{\partial\mathcal{L}}{\partial\Delta} + \frac{\partial\mathcal{L}}{\partial n_2} = 0
\end{equation}
The total joint objective function combines the direct mean squared error fitting loss with these physics constraint losses:
\begin{equation}
\mathcal{L}_{total} = \mathcal{L}_{fit} + 0.2\mathcal{L}_{phys}
\end{equation}
The model was optimized using the Adam optimizer with a learning rate of 0.001, processing batches of 64 samples over 5 epochs. Gradient clipping with a maximum norm of 1.0 was applied to guarantee dynamic training stability.

\section{Evaluation Metrics Definition}
\label{app:metrics}
To quantitatively assess the mathematical fidelity of the inverse mapping, we define the Mean Absolute Error and the Root Mean Square Error over the total dataset points $n$:
\begin{equation}
    \text{MAE} = \frac{1}{n}\sum_{i=1}^{n}|y_{pred, i} - y_{true, i}|
\end{equation}
\begin{equation}
    \text{RMSE} = \sqrt{\frac{1}{n}\sum_{i=1}^{n}(y_{pred, i} - y_{true, i})^2}
\end{equation}

To evaluate the proportion of global variance effectively explained by the algorithmic predictions, we utilize the coefficient of determination:
\begin{equation}
    R^2 = 1 - \frac{\sum_{i=1}^{n}(y_{true, i} - y_{pred, i})^2}{\sum_{i=1}^{n}(y_{true, i} - \bar{y}_{true})^2}
\end{equation}
where $\bar{y}_{true}$ represents the arithmetic mean of the true values. This metric rigorously quantifies the proportion of variance in the dependent physical variable that is reliably predictable from the independent spectral input features.

\begin{figure}[t]
  \centering
  \begin{minipage}{0.48\textwidth}
    \centering
    \includegraphics[width=\linewidth]{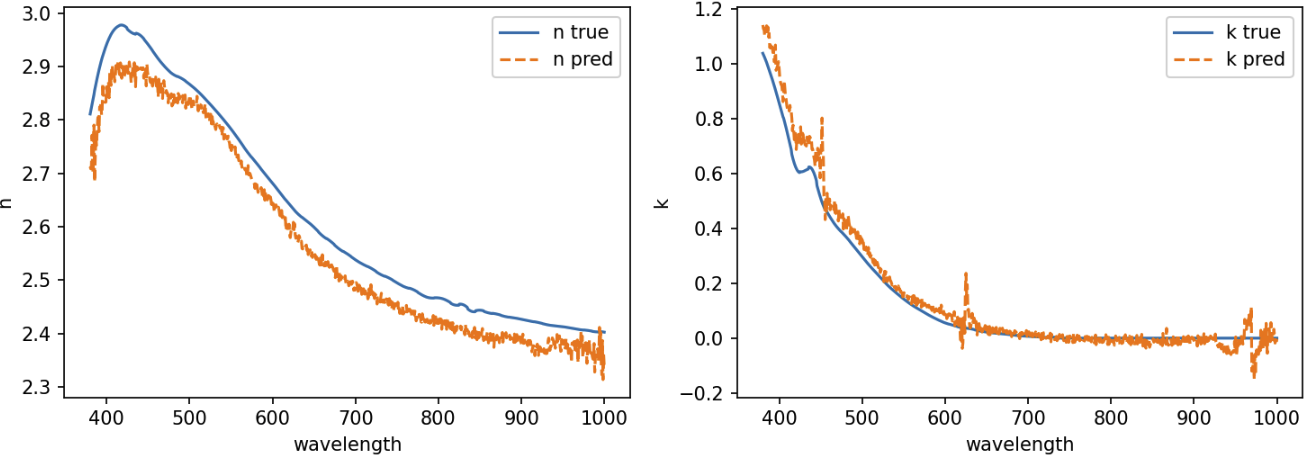}
    \caption*{Indium Oxide ($\text{In}_2\text{O}_3$) at 9.6 nm}
  \end{minipage}\hfill
  \begin{minipage}{0.48\textwidth}
    \centering
    \includegraphics[width=\linewidth]{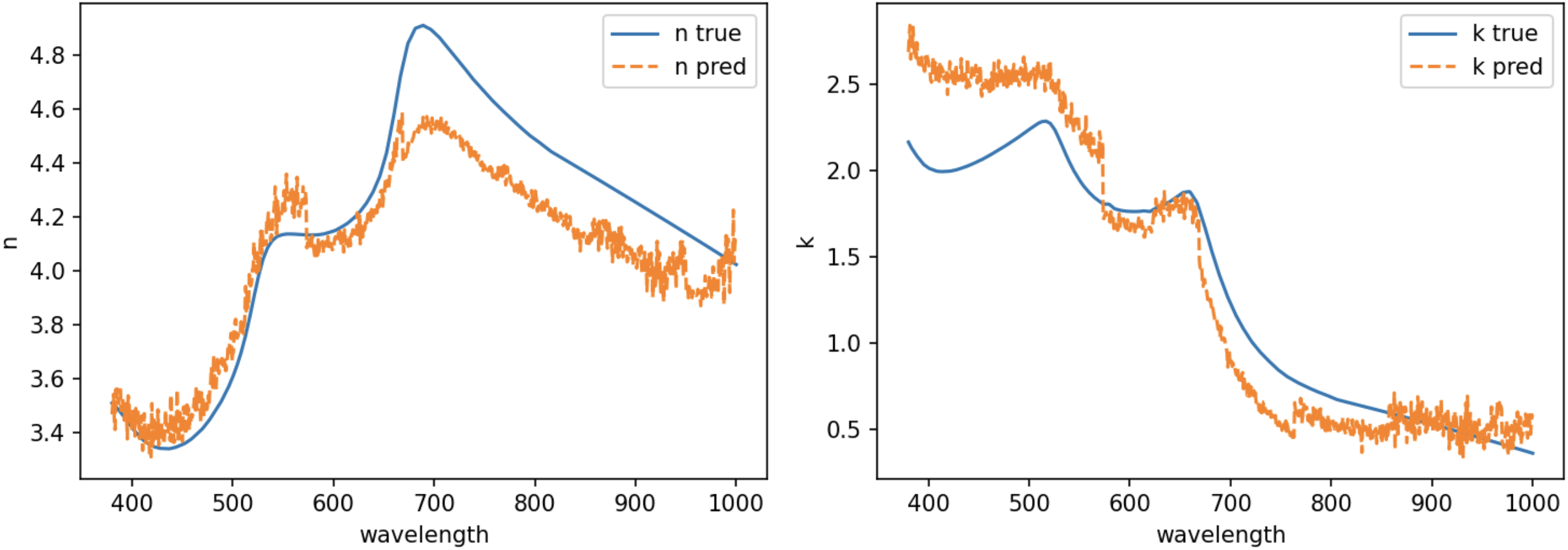}
    \caption*{Indium Antimonide (InSb) at 9.1 nm}
  \end{minipage}
  
  \vspace{0.4cm}
  
  \begin{minipage}{0.48\textwidth}
    \centering
    \includegraphics[width=\linewidth]{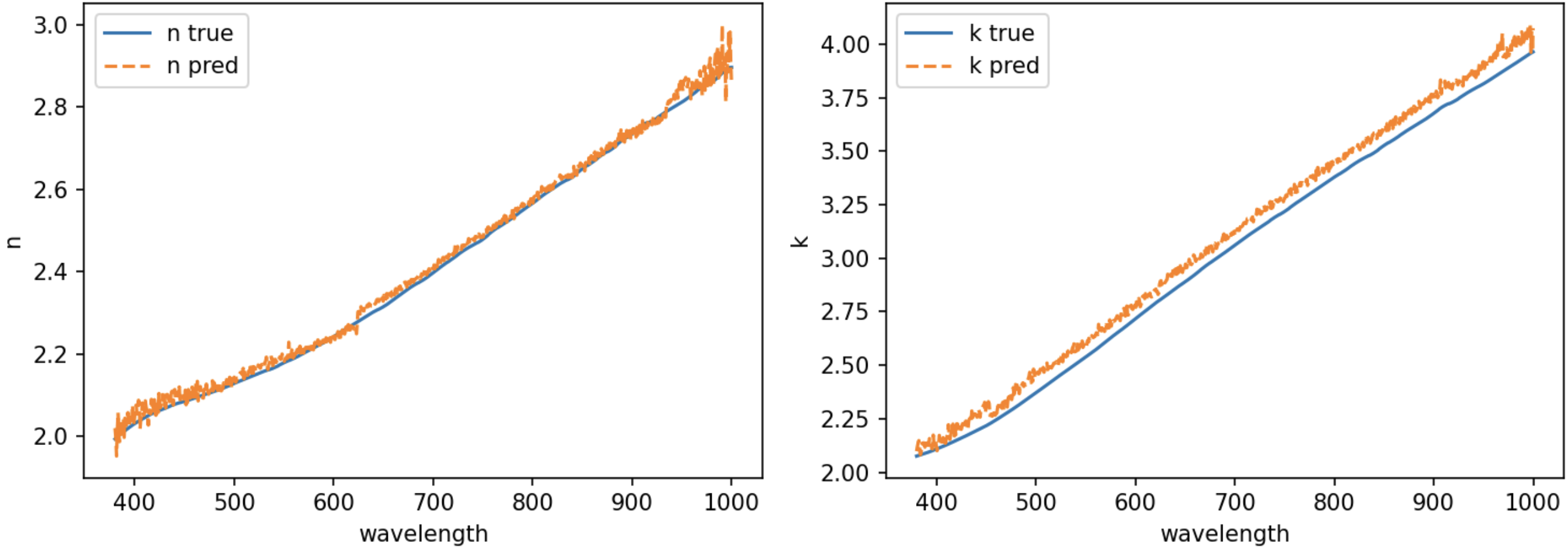}
    \caption*{Nickel Phosphorus (NiP) at 6.6 nm}
  \end{minipage}\hfill
  \begin{minipage}{0.48\textwidth}
    \centering
    \includegraphics[width=\linewidth]{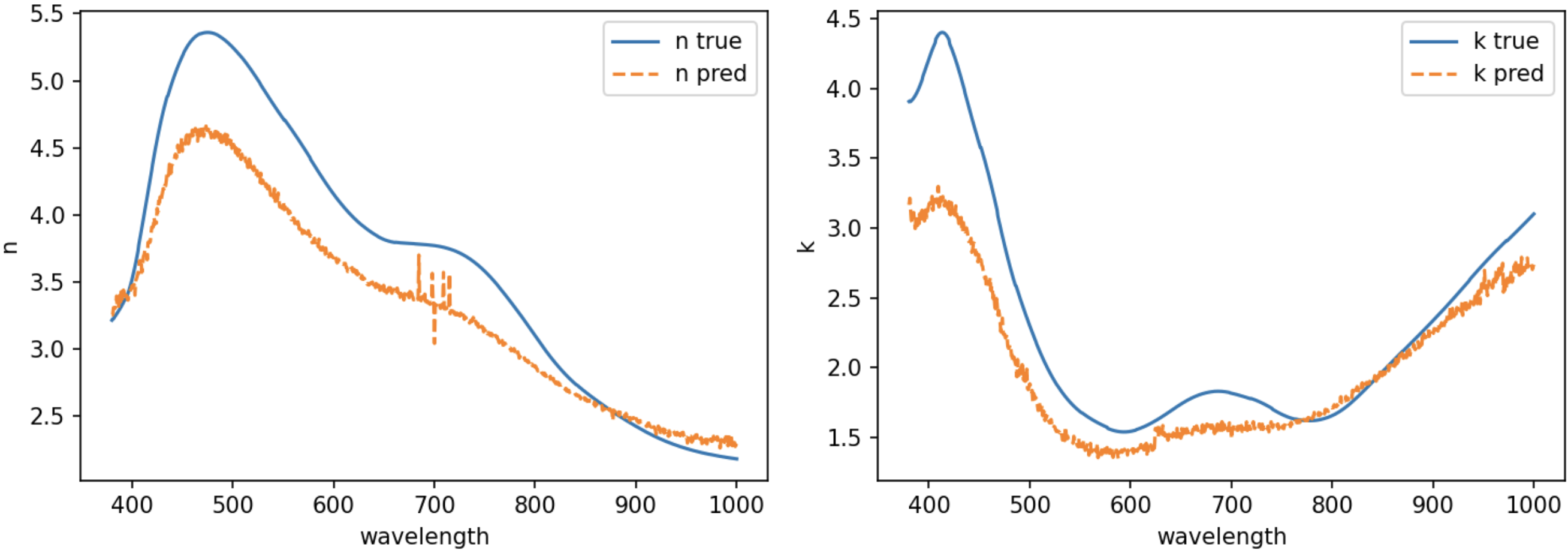}
    \caption*{Osmium (Os) at 1.6 nm}
  \end{minipage}
  \caption{Comparison of the ground truth (solid blue lines) and DCFM predicted (dashed orange lines) optical constants $n$ and $k$ across the 380 to 1000 nm wavelength spectrum. All thin film samples are evaluated on a Strontium Titanate ($\text{SrTiO}_3$) substrate.}
  \label{fig:qualitative_results}
\end{figure}

\section{Qualitative Prediction Results}
\label{app:qualitative}
To visualize the robust generative capacity of the Decoupled Conditional Flow Matching framework, Figure \ref{fig:qualitative_results} presents a direct comparison between the algorithmically predicted and experimentally derived continuous dispersion curves. The optical responses are demonstrated across four distinct material categories utilizing a highly reflective $\text{SrTiO}_3$ substrate. The overlapping spectral trajectories unequivocally validate that the decoupled flow matching mechanism successfully captures extreme localized resonance peaks and complex continuous wavelength dependencies, escaping the unphysical mean averaging trap typical of deterministic regression methodologies.